\documentclass[runningheads]{llncs}
\usepackage{graphicx}

\usepackage{tikz}
\usepackage{comment}
\usepackage{amsmath,amssymb} %
\usepackage{color}
\usepackage{bbm}
\usepackage{graphicx}
\usepackage{booktabs}
\usepackage{xcolor}
\usepackage{multirow}
\usepackage{subcaption}
\usepackage{bbding}
\usepackage{url} %
\usepackage{orcidlink} %
\usepackage{cleveref}

\usepackage[accsupp]{axessibility}  %

\newcommand{\real}{\mathbb{R}}

\newcommand{\figref}[1]{Fig.~\ref{#1}}
\newcommand{\tabref}[1]{Table~\ref{#1}}
\newcommand{\secref}[1]{Sec.~\ref{#1}}
\newcommand{\eqnref}[1]{Eqn.~\ref{#1}}

\begin{document}
\pagestyle{headings}
\mainmatter
\def\ECCVSubNumber{4288}  %

\title{Monitored Distillation %
for Positive Congruent Depth Completion} %

\titlerunning{Monitored Distillation for Positive Congruent Depth Completion}
\author{Tian Yu Liu\inst{1}$^\star$$\orcidlink{0000-0002-4834-051X}$\index{Liu, Tian Yu} \and
Parth Agrawal\inst{1
}$^\star$$\orcidlink{0000-0001-5837-1101}$ \and
Allison Chen\inst{1}\thanks{denotes equal contribution.}$\orcidlink{0000-0002-0919-1849}$ \and \\
Byung-Woo Hong\inst{2}$\orcidlink{0000-0003-2752-3939}$ \and
Alex Wong\inst{3}$\orcidlink{0000-0002-3157-6016}$
}
\authorrunning{T.Y. Liu et al.}
\institute{UCLA Vision Lab, Los Angeles, CA 90095, USA \\
\email{\{tianyu139,parthagrawal24,allisonchen2\}@ucla.edu} \and
Chung-Ang University, Heukseok-Dong, Dongjak-Gu, Seoul, 06973, Korea \\ 
\email{hong@cau.ac.kr} \and
Yale University, New Haven, CT 06511, USA \\
\email{alex.wong@yale.edu} \\
}
\maketitle
\def\method{Monitored Distillation}

\begin{abstract}
    We propose a method to infer a dense depth map from a single image, its calibration, and the associated sparse point cloud. In order to leverage existing models (teachers) that produce putative depth maps, we propose an adaptive knowledge distillation approach that yields a positive congruent training process, wherein a student model avoids learning the error modes of the teachers. In the absence of ground truth for model selection and training, our method, termed \textit{Monitored Distillation}, allows a student to exploit a blind ensemble of teachers by selectively learning from predictions that best minimize the reconstruction error for a given image. Monitored Distillation yields a distilled depth map and a confidence map, or ``monitor'', for how well a prediction from a particular teacher fits the observed image. The monitor adaptively weights the distilled depth where if all of the teachers exhibit high residuals, the standard unsupervised image reconstruction loss takes over as the supervisory signal. On indoor scenes (VOID), we outperform blind ensembling baselines by 17.53\%
    and unsupervised methods by 24.25\%; we boast a 79\% model size reduction while maintaining comparable performance to the best supervised method. For outdoors (KITTI), we tie for 5th overall on the benchmark despite not using ground truth.
    Code available at: \url{https://github.com/alexklwong/mondi-python}.
\end{abstract}

\section{Introduction}
Interaction with physical space requires a representation of the 3-dimensional (3D) geometry of the surrounding environment. Most mobile platforms include at least one camera and some means of estimating range at a sparse set of points i.e. a point cloud. These could be from a dedicated range sensor such as a LiDAR or radar, or by processing the images using a visual odometry module. Depth completion consists of inferring a dense depth map, with a range value corresponding to every pixel, from an image and a sparse point cloud. Inherently, depth completion is an ill-posed inverse problem, so priors need to be imposed in the form of generic regularization or learned inductive biases.

Natural scenes exhibit regularities that can be captured by a trained model, for instance a deep neural network (DNN), using a dataset of images and corresponding sparse depths. While we wish to avoid any form of manual or ground truth supervision, we also strive to exploit the availability of differing types of pretrained models, whether from synthetic data or other supervised or unsupervised methods. We refer to these pretrained models as ``teachers,'' each providing a hypothesis of depth map for a given image and sparse point cloud. This leads to a blind ensemble setting where ground truth is not available (e.g. transferring models trained on a specific task to new datasets with no ground truth) for the explicit evaluation of pretrained models i.e. model selection. The key question, then, is how to make use of a heterogeneous collection of teachers, along with other variational principles such as minimization of the photometric reprojection error and generic regularizers such as structural similarity. 

In general, different teachers will behave differently not only
across images, but even across regions within a given image.
The incongruency of different models trained on the same tasks has been observed in the context of classification model versioning \cite{yan2021positive}. Particularly, the same architecture trained with the same data, but starting from different initial conditions can yield models that \textit{differ on a significant portion of the samples} while achieving the same average error rate. Thus, a naive ensembling of a handful of teachers yields the union of the failure modes, only modestly mitigated by the averaging.

Instead, we propose \method\ for selecting which teacher to emulate in each image at each pixel. The selection is guided by a ``monitor'',  based on the residual between the observations (e.g. image, sparse point cloud) and their reconstructions generated by each teacher. This yields a spatially-varying confidence map that weights the contribution of the selected teachers as well as the structural and photometric reprojection errors i.e. unsupervised losses, customary in structure-from-motion. In doing so, our method is robust even when poor performing teachers are introduced into the ensemble -- discarding their hypotheses in favor of the ones that better reconstruct the scene. In the extreme case where every teacher produces erroneous outputs, our method would still learn a valid depth estimate because of our unsupervised fall-back loss.

\textbf{Our contributions} are as follows: 
\textbf{(i)} We propose an adaptive method to combine the predictions of a blind ensemble of teachers based on their compatibility with the observed data; to the best of our knowledge, we are the first to propose knowledge distillation from a blind ensemble for depth completion. 
\textbf{(ii)} The adaptive mechanism yields a spatially varying confidence map or ``monitor'' that modulates the contributions of each teacher based on their residuals, leading to a training method that is positive congruent. 
\textbf{(iii)} Even when all members of the ensemble fail, our model automatically reverts to the unsupervised learning criteria and generic regularization, allowing us to avoid distilling erroneous knowledge from teachers.
\textbf{(iv)} Our method outperforms distillation and unsupervised methods by 17.53\% and 24.25\%
respectively on indoors scenes; we are comparable to top supervised methods with a 79\% model size reduction. On the KITTI benchmark, we tie for 5th overall despite not using ground truth.

\section{Related Works}
Depth completion is a form of imputation and thus requires regularization, which may come from generic assumptions or learned from data. The question is: How to best combine different sources of regularization, adaptively \cite{hong2017adaptive,hong2019adaptive,wong2019bilateral}, in a way that leverages their strengths, while addressing their weaknesses?

\textit{Supervised depth completion} is trained by minimizing a loss with respect to ground truth. Early methods posed the task as learning morphological operators \cite{dimitrievski2018learning} and compressive sensing \cite{chodosh2018deep}. Recent works focus on network operations \cite{eldesokey2018propagating,huang2019hms} and design \cite{chen2019learning,ma2019self,uhrig2017sparsity,yang2019dense} to effectively deal with the sparse inputs. 
For example, \cite{li2020multi} used a cascade hourglass network, \cite{jaritz2018sparse,yang2019dense} used separate image and depth encoders and fused their representations, and \cite{huang2019hms} proposed an upsampling layer and joint concatenation and convolution. Whereas, \cite{eldesokey2020uncertainty,eldesokey2018propagating,qu2021bayesian,qu2020depth} learned uncertainty of estimates, \cite{van2019sparse} leveraged confidence maps to fuse predictions from different modalities, and \cite{qiu2019deeplidar,xu2019depth,zhang2018deep} used surface normals for guidance. 
\cite{cheng2020cspn++,park2020non} use convolutional spatial propagation networks, \cite{hu2021penet} used separate image and depth networks and fused them with spatial propagation. 
While supervised methods currently hold the top ranks on benchmark datasets i.e. KITTI \cite{uhrig2017sparsity} and VOID \cite{wong2020unsupervised}, they inevitably require  ground truth for supervision, which is typically unavailable. Furthermore, these architectures are often complex and require many parameters (e.g. 132M for \cite{hu2021penet}, 53.4M for \cite{qiu2019deeplidar}, and 25.8M for \cite{park2020non}), making them computationally prohibitive to train and impractical to deploy \cite{merrill2021robust}.

\textit{Unsupervised depth completion} assumes that additional data (stereo or monocular videos) is available during training. Both stereo \cite{shivakumar2019dfusenet,yang2019dense} and monocular \cite{ma2019self,wong2021learning,wong2021adaptive,wong2020unsupervised} paradigms focus largely on designing losses that minimize (i) the photometric error between the input image and its reconstructions from other views, and (ii) the difference between the prediction and sparse depth input (sparse depth reconstruction). Architecture-wise, \cite{wong2021unsupervised} proposed a calibrated backprojection network. However, all of these methods rely on \textit{generic} regularization i.e. local smoothness that is not informed by the data. Attempts to leverage learned priors mainly focused on synthetic data. \cite{lopez2020project} applied image translation to obtain ground truth in the real domain; whereas \cite{wong2020unsupervised,yang2019dense} used synthetic data to learn a prior on the shapes populating a scene. 
We also employ an unsupervised loss, but unlike them, we distill regularities from a blind ensemble of pretrained models that can be trained on synthetic or real data, supervised or unsupervised.

\textit{Knowledge Distillation} uses a simpler student model to approximate the function learned by a larger, more complex teacher model by training it to learn the soft target distribution \cite{hinton2015distilling}. There exists many works on knowledge distillation, including image classification \cite{liu2018multi,romero2014fitnets,xiang2020learning}, object detection \cite{chawla2021data,chen2017learning,chen2019new}, semantic segmentation \cite{liu2019structured,michieli2021knowledge,park2020knowledge}, depth estimation \cite{hu2021boosting,liu2020structured,wang2021knowledge}, and more recently, depth completion \cite{hwang2021lidar}. \cite{choi2021stereo,liu2019structured,liu2020structured} utilize pairwise and holistic distillation to capture structural relationships and \cite{michieli2021knowledge} distills latent representations to guide learning. \cite{wang2021knowledge} leverages knowledge distillation for monocular depth estimation on mobile devices, and \cite{pilzer2019refine} uses cyclic inconsistency and knowledge distillation for unsupervised depth estimation, where the student network is a sub-network of the teacher.
In depth completion, \cite{hwang2021lidar} uses knowledge distillation for joint training of both teacher and student models. Unlike ours, this method uses ground truth.

Ensemble learning addresses the limitations of a single teacher by distilling information from multiple teachers \cite{fukuda2017efficient}. If done effectively, the student will learn to extract the most relevant information from each teacher. This has been explored in classification \cite{kang2020ensemble,lan2018knowledge,walawalkar2020online}, but fewer works utilize it in dense prediction tasks. \cite{chao2021rethinking} uses it for domain adaptation in semantic segmentation and \cite{gofer2021adaptive} in selecting lidar points for depth completion. We further assume the blind ensemble setting \cite{traganitis2018blind} where we lack ground truth for evaluation of the ensemble.

\textit{Positive congruent training} \cite{yan2021positive} observed sample-wise inconsistencies in classification versioning, where new models wrongly predict for samples that were previously classified correctly by an older, reference model on the same task and dataset. To address this, they propose to emulate the reference model (teacher) only when its predictions are correct; otherwise, they minimize a loss with respect to ground truth -- yielding reduced error rates and inconsistencies.
Monitored distillation is inspired by positive-congruency, but unlike \cite{yan2021positive}, we do not require ground truth and are applicable towards geometric tasks.

\section{Method Formulation}
We wish to recover the 3D scene from a calibrated RGB image $I : \Omega \subset \real^2 \mapsto \real^3_+$ and its associated sparse point cloud projected onto the image plane $z : \Omega_{z} \subset \Omega \mapsto \real_+$. To do so, we propose learning a function $f_\theta$ that takes as input $I, z$, and camera intrinsics $K$ and outputs a dense depth map $\hat{d} := f_\theta (I, z, K) \in \real_+^{H \times W}$. 

\begin{figure}[t]
\centering
\includegraphics[width=0.99\linewidth]{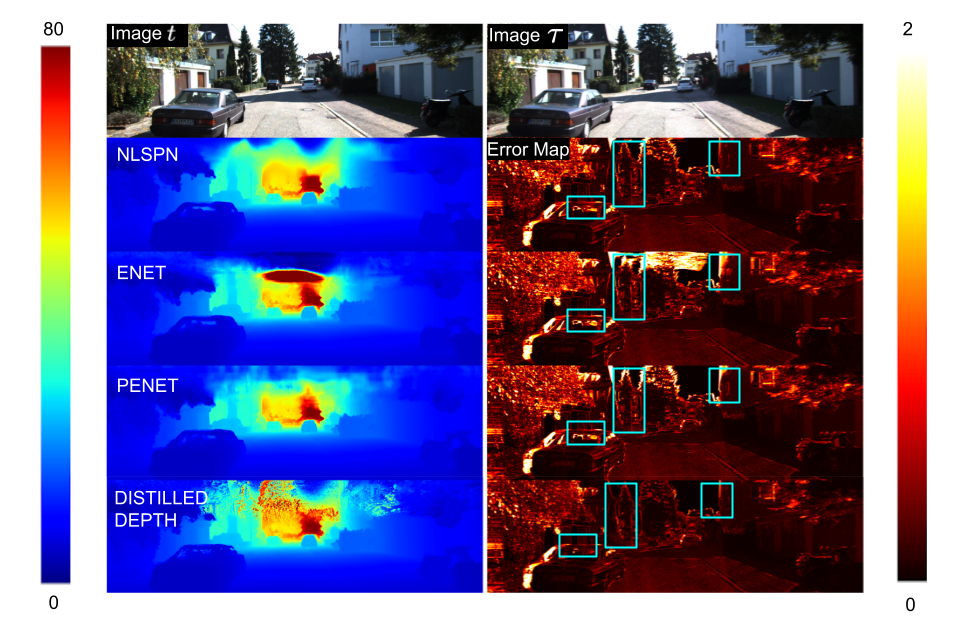}
\caption{
    \textbf{Error Modes of Teacher Models on KITTI}. Row 1 shows the input image $I_t$ (left) and an image taken from another view $I_\tau$ (right). Rows 2-4 shows the predicted depth maps (left) and error maps (right) for each teacher, where each has different error modes. Row 5 shows
    the distilled depth that our method adaptively constructs from the teacher models. The error for each region in the distilled depth lower bounds the reprojection error of the individual teachers.
} 
\label{fig:teacher-error-modes}
\end{figure}

We assume that for each synchronized pair of image and sparse depth map $(I_t, z_t)$ captured at a viewpoint $t$, we have access to a set of spatially and/or temporally adjacent alternate views $T$ and the corresponding set of images $I_T$.
Additionally, we assume access to a set of $M$ models or ``teachers'' $\{ h_{i} \}_{i=1}^{M}$ (e.g. publicly available pretrained models). \figref{fig:teacher-error-modes} shows that each teacher has unique failure modes.
As we operate in the blind ensemble setting, we lack ground truth to evaluate teacher performance for model selection. To address this, we propose \method, an adaptive knowledge distillation framework for ensemble learning that results in positive congruent training: We only learn from a teacher if its predictions are compatible with the observed scene. 

\begin{figure*}[t]
    \centering
    \includegraphics[width=\linewidth]{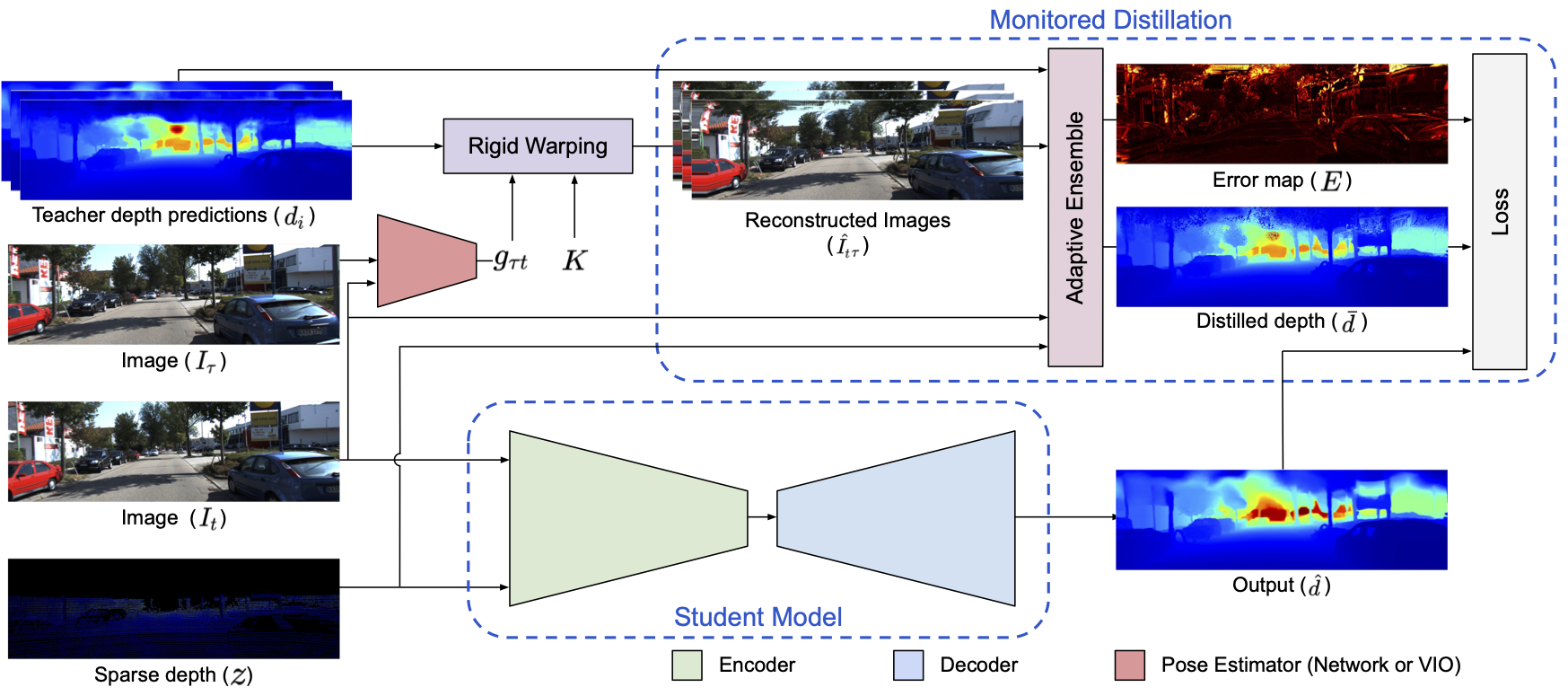}
    \caption{\textbf{Monitored Distillation.} Our method measures the reconstruction residual of predictions from each teacher and constructs the distilled depth $\bar{d}$ based on a pixel-wise selection of predictions that best minimize the reconstruction error $E$. We derive a monitor function $Q$ from $E$, which adaptively balances the trade-offs between the distilling from the ensemble and the unsupervised losses.
    } 
\label{fig:system-diagram}
\end{figure*}

\begin{figure}[t]
    \centering
    \begin{subfigure}{0.38\textwidth}
    \includegraphics[width=\linewidth]{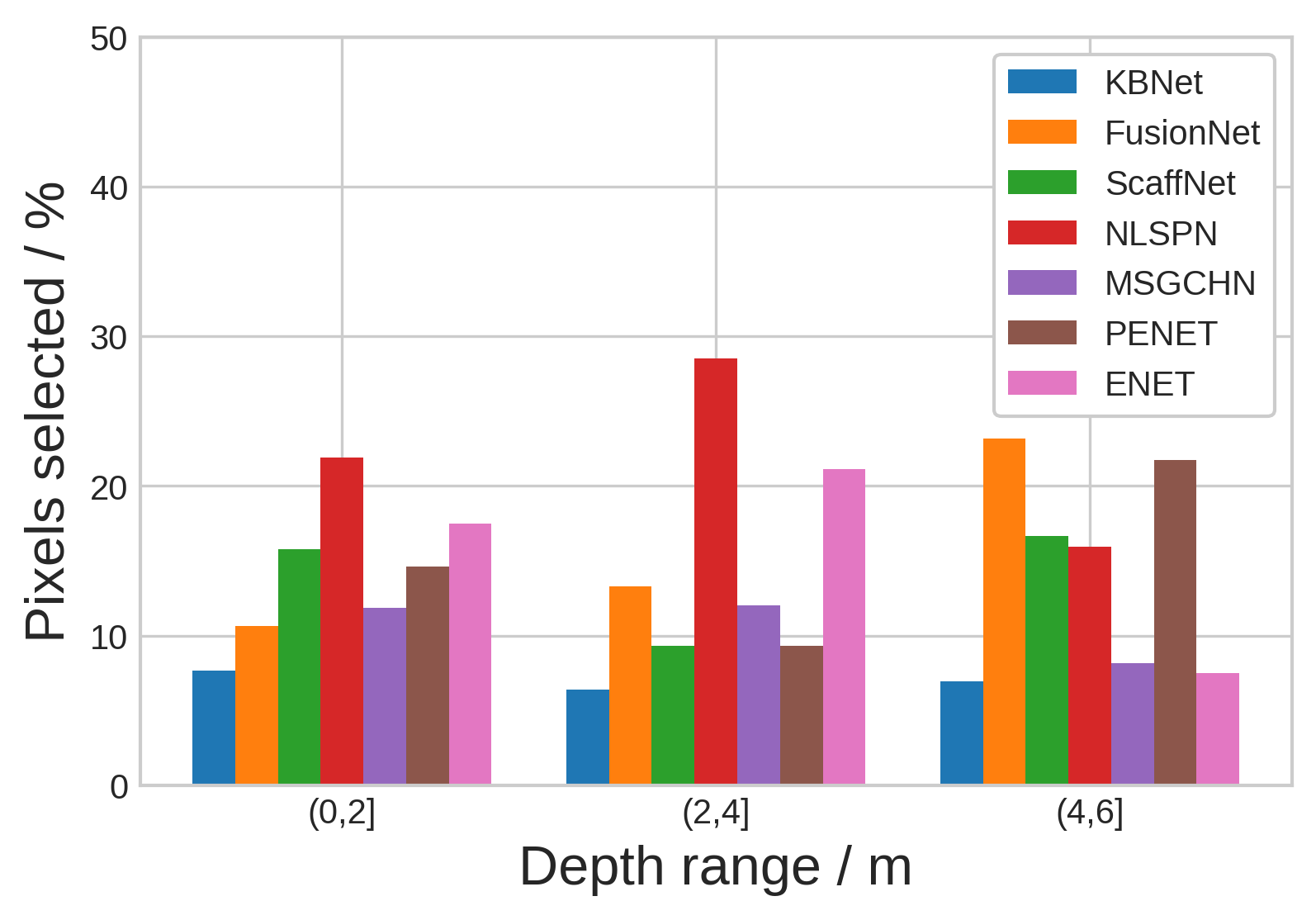}
    \caption{VOID Dataset (indoor)}
    \end{subfigure}
\begin{subfigure}{0.4\textwidth}
    \includegraphics[width=\linewidth]{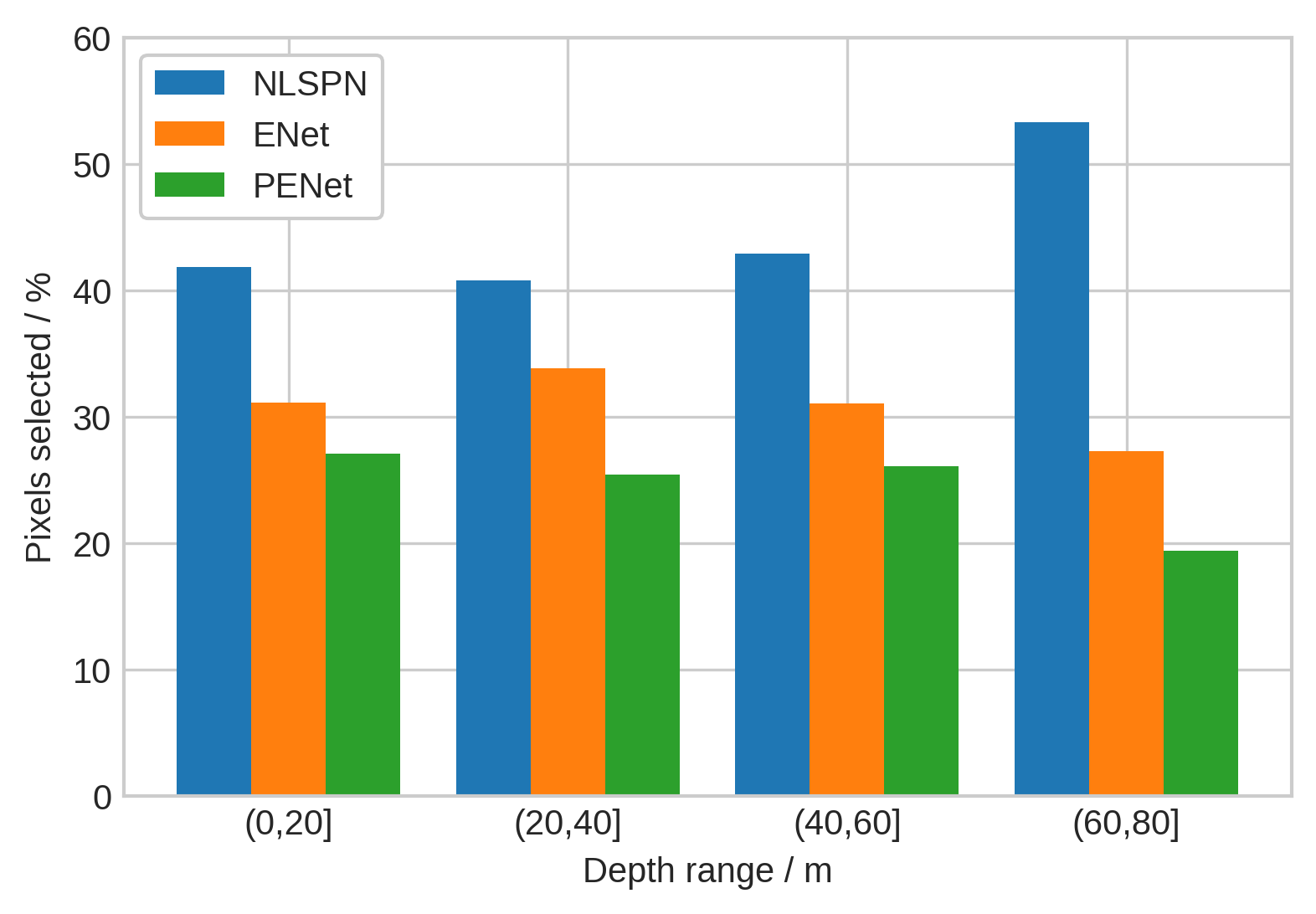}
    \caption{KITTI Dataset (outdoor)}
    \end{subfigure}
    \caption{
        \textbf{Teacher Selection Distribution.} The plots show the proportion of pixels selected from each teacher model. Note: error modes vary across different depth ranges as different teachers dominate the selection at different distances.
    } 
\label{fig:teacher-distribution}
\end{figure}
To this end, we leverage geometric constraints between $I_t$ and $I_\tau \in I_T$ and validate the correctness of predictions $d_i := h_{i}(I_t, z_t)$ produced by each teacher through averaging their photometric reprojection residuals from different views $I_T$ and weighting them based on deviations from $z$. 
From the error, we derive a confidence map that determines the compatibility of each teacher to the observed image $I_t$. 
We then construct distilled depth $\bar{d} \in \real^{H \times W}_+$ via pixel-wise selection from the ensemble that yields the highest confidence.
The resulting spatially varying confidence map acts as a ``monitor" to balance the trade-off between ``trusting'' the ensemble and falling back onto unsupervised geometric consistency as a supervisory signal (i.e. when all teachers yield high residuals). \\

\noindent\textbf{Monitored Distillation.}
Given $M \in \mathbb{Z}^{+}$ teachers and their predicted depth maps $d_{i}$, for $i \in \{ 1,\cdots, M \}$, we construct a distilled depth map $\bar{d}$ by adaptively selecting predictions from the teacher ensemble that best minimize reconstruction error of the observed point cloud and image. To this end, we reconstruct the observed image $I_t$ via reprojection from an adjacent view $I_\tau, \tau \in T$:
\begin{equation}
    \hat{I}_{t \tau}(x, d) = I_\tau ( \pi g_{\tau t} K^{-1} \bar{x} d(x)),
\label{eqn:reprojection}
\end{equation}
where $d$ denotes the depth values for $x \in \Omega$, $\bar{x} = [x^\top \ 1]^\top$ is the homogeneous coordinate of $x$, $g_{\tau t} \in SE(3)$ is the relative pose of the camera from view $t$ to $\tau$, $K$ is the camera intrinsics, and $\pi$ is the perspective projection. In practice, $g_{\tau t}$ can be derived from camera baseline if $I_t$ and $I_\tau$ are stereo pairs, directly estimated by a visual inertial odometry (VIO) system, or learned from a pose network if the views are taken from a video.

For each teacher $h_i$, we measure the photometric reprojection error $P_i$ via the mean SSIM \cite{wang2004image} between $I_t$ and each reconstruction $\hat{I}_{t \tau}(x, d_i), \tau \in T$:
\begin{equation}
    P_i(x) = 
  	     \frac{1}{|T|}\sum_{\tau \in T}\big(1 - \phi(\hat{I}_{t \tau}(x, d_i), I_{t}(x))\big)
\label{eqn:reproj_error}
\end{equation}
For ease of notation, we denote SSIM as $\phi(\cdot)$. 

As photometric reprojection alone does not afford scale, we additionally measure the local deviation of teacher predictions from the observed sparse point cloud within a $k \times k$ neighborhood of $x$, denoted by $\mathcal{N}(x)$:
\begin{equation}
    Z_i = \frac{1}{k^2|z_t|} \sum_{x \in \Omega} \sum_{y \in \mathcal{N}(x)} \mathbbm{1}_{z_t(x)} \cdot |d_i(y) - z_t(x)|
\end{equation}
When used as a weight, $\beta_i := 1 - \exp(-\alpha Z_i)$ serves to resolve the scale ambiguity between different teachers, where $\alpha$ is a temperature parameter. We can then define $E_i$, the weighted reconstruction residual from the $i$-th teacher, as:
\begin{equation}
    E_i(x) = \beta_i P_i(x).
    \label{eqn:error_map}
\end{equation} 
To construct the distilled depth, we selectively choose the depth prediction for each pixel $x \in \Omega$ that minimizes the overall residual error $E_i$ across all teachers: 
\begin{equation}
    \bar{d}(x) = \sum_{i=1}^{M} \mathbbm{1}_i(x) d_i(x),
\end{equation}
where $\mathbbm{1}_i$ is a binary weight map of the $i$-th teacher is given by
\begin{align}
    \mathbbm{1}_i(x) = \begin{cases}
        1 & E_i(x) < E_j(x) \ \forall \ j \neq i  \\
        0 &\text{otherwise.} \\
    \end{cases}
\end{align}
In other words, $\mathbbm{1}_i(x) = 1$ when $d_i(x)$ yields the lowest photometric residual. \figref{fig:system-diagram} shows an overview of our method, where teacher predictions are ensembled into distilled depth for supervision. \figref{fig:teacher-distribution} shows the distribution of teachers chosen for constructing the distilled depth. As observed, different teachers perform well in different regions across different depth ranges. Our method selects points from each teacher with the lowest error (i.e. highest confidence) to yield an adaptive ensemble (see \figref{fig:teacher-error-modes}). 

Despite being trained on ground truth, each teacher can only \textit{approximate} the true distribution of depths in the scene.
While selectively ensembling based on the reprojection residual will address \textit{some} error modes of the teachers, it is still possible for all teachers to yield high reconstruction residuals. 
Hence, we do not trust the ensemble fully, and instead further adaptively weight the ensemble supervision with a monitor $Q$ based on the error of the distilled depth $\bar{d}$. 
As we have already constructed the error maps $E_i$ for each teacher,
we can similarly aggregate the error for each pixel $E(x) = \min_i E_i(x)$ for $x \in \Omega$.
The final monitor $Q \in [0, 1]^{H \times W}$
is a spatially adaptive per-pixel confidence map:
\begin{equation}
    Q(x) = \exp (- \lambda E(x)),
\end{equation}
where $\lambda$ is a temperature parameter (see Supp. Mat.). 
$Q$ naturally assigns higher confidence to points in the distilled depth that are compatible with the observed image $I_t$ as measured by reconstruction error, and is used to weight the supervision signal. Our monitored knowledge distillation objective reads:
\begin{equation}
    \ell_{md} = \frac{1}{|\Omega|} \sum_{x \in \Omega} Q(x) \cdot |\hat{d}(x) - \bar{d}(x)|.
\end{equation}

Typically, a student learns the error modes of its teacher. But by distilling from the adaptive ensemble of teachers that is positive congruent, our student model learns not to make the same mistake as any individual teacher. We refer to this process as \textit{\method}, in which our monitor function $Q$ gives higher weight to the teachers in regions of lower reconstruction error. For regions where all of the teachers within the ensemble yield high residuals, we default to unsupervised loss to avoid learning the common error modes of any teacher. \\

\noindent\textbf{Unsupervised Objective.}
For regions with high reconstruction error, the monitoring function $Q$ allows us to fall back onto standard unsupervised photometric reprojection error, i.e. color and structural consistencies, as the training signal:
\begin{equation}
     \ell_{co} = \frac{1}{|\Omega|}\frac{1}{|T|} \sum_{x \in \Omega}\sum_{\tau \in T}  
     (1 - Q(x))\big(|\hat{I}_{t \tau}(x, \hat{d})-I_{t}(x)|\big)
\label{eqn:loss_co}
\end{equation}
\begin{equation}
    \ell_{st} = \frac{1}{|\Omega|}\frac{1}{|T|} \sum_{x \in \Omega}\sum_{\tau \in T}
    (1 - Q(x)) \big(1 -  \phi(\hat{I}_{t \tau}(x, \hat{d}), I_{t}(x))\big)
\label{eqn:loss_st}
\end{equation}
We weight the relative contributions of these losses with the complement of our adaptive monitor function ($1 - Q$). As a result, our framework naturally allows us to search for the correct correspondences (and in turn better depth estimation) in regions where the ensemble failed. In other words, regions which the monitor deems as high confidence are more heavily influenced by $\ell_{md}$ as supervision, while lower confidence regions will minimize unsupervised losses instead.

Because the ensemble is informed by large amounts of data, their predictions have regularities of our physical world, e.g. roads are flat and surfaces are locally connected, ``baked into'' them. This presents an advantage: The student will learn priors, often too complex to be modeled by generic assumptions, from the ensemble. However, these priors may backfire when all the teachers yield high residuals. Luckily, $Q$ naturally limits the influence of the ensemble in such cases, but this in turn reduces the amount of regularization that is needed for ill-posed problems like 3D reconstruction. Hence, for these cases, we default to generic assumptions i.e. a local smoothness regularizer: 
\begin{equation}
    \ell_{sm} = \frac{1}{|\Omega|} \sum_{x \in \Omega}           (1-Q(x))
      	    \big(\lambda_{X}(x)|\partial_{X}\hat{d}(x)|+
      	    \lambda_{Y}(x)|\partial_{Y}\hat{d}(x)|\big)
\end{equation}
where $\partial_{X}, \partial_{Y}$ are gradients along the x and y directions, weighted by $\lambda_{X} := e^{-|\partial_{X}I_{t}(x)|}$ and $\lambda_{Y} := e^{-|\partial_{Y}I_{t}(x)|}$ respectively.

Thus, we have the following overall loss function
\begin{equation}
    \mathcal{L} = w_{md} \ell_{md}  + w_{ph} \ell_{ph} + w_{st} \ell_{st} + w_{sm} \ell_{sm}
\label{eqn:loss}
\end{equation}
where $w_{(\cdot)}$ denotes the respective weights for each loss term (see Supp. Mat.). \\

\noindent\textbf{Student Model Architecture.}
Through monitored distillation from an ensemble of teachers, a simpler student model can be trained on the output distribution of more complex teacher models to achieve comparable performance. 
Therefore, we compress KBNet \cite{wong2021unsupervised} by replacing the final two layers in the encoder with depth-wise separable convolutions \cite{howard2017mobilenets} to yield a $23.2\%$ reduction in the number of model parameters. Compared to the best supervised teacher models that require 25.84M (NLSPN \cite{park2020non}), 131.7M (ENet \cite{hu2021penet}), 132M (PENet \cite{hu2021penet}), and 6.9M (the original KBNet) parameters, our student model only requires 5.3M.

\begin{table}[t]
    \scriptsize
    \centering
    \caption{\textbf{Blind Ensemble Distillation.} We compare Monitored Distillation against naive ensembling methods for training a student model.
    }
    \setlength\tabcolsep{8pt}
    \begin{tabular}{l l c c c c c c}
        \midrule 
        Ensemble Type & Distillation Method & MAE & RMSE & iMAE & iRMSE \\ 
        \midrule
        \multirow{1}{*}{None}
        & Unsupervised Loss Only
        & 55.67 & 117.21 & 28.68 & 58.31 \\
        \midrule
        \multirow{5}{*}{Supervised}
        & Mean w/o Unsupervised Loss
        & 34.27 & 91.72 & 17.63 & 41.39 \\
        & Mean
        & 34.04 & 89.19 & 17.30 & 40.43 \\
        & Median
        & 34.64 & 89.80 & 17.46 & 39.77 \\
        & Random
        & 35.18 & 92.30 & 18.41 & 42.95 \\
        & Ours w/o $\beta$
        & 32.86 & 85.53 & 16.44 & 39.14 \\
        & \textbf{Ours}
        & \textbf{30.88} & \textbf{87.48} & \textbf{15.31} & \textbf{38.33} \\
        \midrule
        \multirow{6}{*}{Unsupervised}
        & Mean w/o Unsupervised Loss
        & 44.73 & 96.56 & 24.08 & 49.55 \\
        & Mean 
        & 41.96 & 94.47 & 23.80 & 50.37 \\
        & Median
        & 43.86 & 99.46 & 23.62 & 50.85 \\
        & Random
        & 39.38 & 92.14 & 20.62 & 46.04 \\
        & Ours w/o $\beta$
        & 38.78 & 90.72 & 20.53 & 45.91 \\
        & \textbf{Ours}
        & \textbf{36.42} & \textbf{87.78} & \textbf{19.18} & \textbf{43.83} \\
        \midrule
        \multirow{6}{*}{Heterogeneous}
        & Mean w/o Unsupervised Loss
        & 44.53 & 100.59 & 23.33 & 48.36 \\
        & Mean
        & 35.79 & 84.78 & 18.65 & 42.90 \\
        & Median
        & 33.89 & 85.25 & 17.31 & 40.40 \\
        & Random
        & 43.64 & 94.38 & 24.74 & 50.27  \\
        & Ours w/o $\beta$
        & 32.09 & 80.20 & 16.15 & 38.86 \\
        & \textbf{Ours}
        & \textbf{29.67} & \textbf{79.78} & \textbf{14.84} & \textbf{37.88} \\
        \midrule
    \end{tabular}
\label{tab:master_comparisons}
\end{table}

\section{Experiments}
\label{sec:experiments}

We evaluate our method on public benchmarks -- VOID \cite{wong2020unsupervised} for indoor and outdoor scenes and KITTI \cite{uhrig2017sparsity} for outdoor driving settings. We describe evaluation metrics, implementation details, hyper-parameters and learning schedule in the Supp. Mat. \cite{liu2022monitored}. All experiments are performed under the blind ensemble setting where we do not have ground truth for model selection nor training.
 
\textit{VOID dataset} \cite{wong2020unsupervised} contains synchronized $640 \times 480$ RGB images and sparse depth maps of indoor (laboratories, classrooms) and outdoor (gardens) scenes. The associated sparse depth maps contain $\approx$1500 sparse depth points with a density of  $\approx$0.5\%. They are obtained by a set of features tracked by XIVO \cite{fei2019geo}, a VIO system. The dense ground-truth depth maps are acquired by active stereo. As opposed to static scenes in KITTI, the VOID dataset contains 56 sequences with challenging motion. Of the 56 sequences, 48 sequences ($\approx$45,000 frames) are designated for training and 8 for testing ($800$ frames). We follow the evaluation protocol of \cite{wong2020unsupervised} and cap the depths between 0.2 and 5.0 meters.

\textit{KITTI dataset} \cite{uhrig2017sparsity} depth completion benchmark contains $\approx$86,000 raw $1242 \times 375$ image frames (43K stereo pairs) and synchronized sparse depth maps . The sparse depth is obtained using a Velodyne lidar sensor and, when projected, covers $\approx$ 5\% of the image space. The ground truth depths are semi-dense, which we use only for evaluation purposes. We use the designated 1,000 samples for validation and evaluate test-time accuracy on KITTI's online testing server. 

\textit{Teacher ensembles:} We use the following ensembles for VOID (\tabref{tab:master_comparisons}, \ref{tab:teacher_multiple_domains}, \ref{tab:void_test_set_results}): (i) supervised ensemble of NLSPN \cite{park2020non}, MSG-CHN \cite{li2020multi}, ENet, and PENet \cite{hu2021penet}, (ii) unsupervised ensemble of FusionNet \cite{wong2021learning}, KBNet \cite{wong2021unsupervised}, and ScaffNet \cite{wong2021learning} (trained on SceneNet \cite{mccormac2017scenenet}), and (iii) heterogeneous ensemble of all seven methods.
For KITTI (\tabref{tab:kitti_unsupervised_benchmark}, \ref{tab:kitti_supervised_benchmark}), we used NLSPN \cite{park2020non}, ENet, and PENet \cite{hu2021penet}. \\

\begin{table}[t]
    \scriptsize
    \centering
    \caption{\textbf{Different Teacher Ensembles.} We apply \method\ to various combinations of teachers trained on different datasets. Using an ensemble trained only on NYUv2($\ddagger$) and SceneNet($\dagger$) still benefits a student on VOID($\diamond$). 
    }
    \setlength\tabcolsep{6pt}
    \begin{tabular}{llcccccc}
        \midrule 
        Teachers & Teachers Trained On & MAE & RMSE & iMAE & iRMSE \\ 
        \midrule
        None (Unsupervised Loss Only) & - & 55.67 & 117.21 & 28.68 & 58.31 \\
        FusionNet$^\diamond$, ScaffNet$^\dagger$ & VOID, SceneNet & 48.72 & 102.44 & 26.94 & 56.32 \\
        FusionNet$^\diamond$, KBNet$^\diamond$ & VOID & 40.10 & 92.03 & 22.16 & 46.86  \\
        KBNet$^\diamond$, ScaffNet$^\dagger$ & VOID, SceneNet & 38.87 & 91.76 & 20.50 & 46.67 \\
        FusionNet$^\diamond$, KBNet$^\diamond$, ScaffNet$^\dagger$ & VOID, SceneNet & 36.42 & 87.78 & 19.18 & 43.83 \\
        FusionNet$^\ddagger$, KBNet$^\ddagger$, ScaffNet$^\dagger$ & NYUv2, SceneNet  & 46.66 & 104.05 &    26.13 & 54.96\\
        \midrule
    \end{tabular}
\label{tab:teacher_multiple_domains}
\end{table}

\begin{figure}[t]
    \centering
    \includegraphics[width=1.0\linewidth]{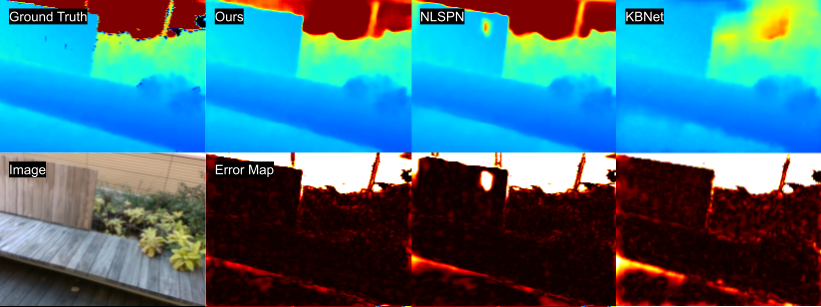}
        \caption{
            \textbf{\method\ vs. Supervised and Unsupervised Teachers}. We address the failure modes in the top supervised method NLSPN \cite{park2020non} (hole in the bench) by distilling from a heterogeneous ensemble.
        } 
\label{fig:void_comparison}
\end{figure}

\begin{figure}[t]
    \centering
    \includegraphics[width=1.0\linewidth]{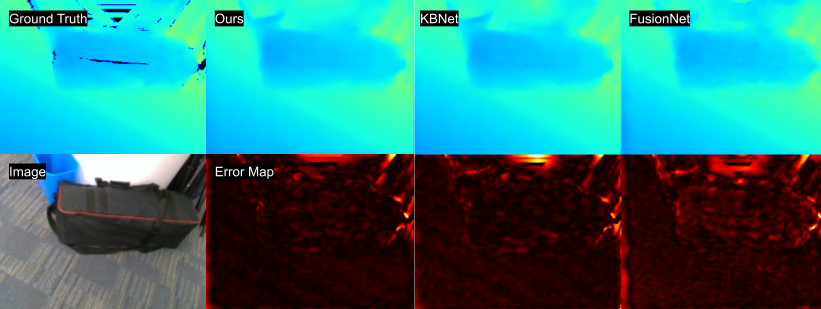}
        \caption{
            \textbf{\method\ vs. Unsupervised Teachers}. 
            While KBNet \cite{wong2021unsupervised} is the best performer among unsupervised methods, by ensembling it with weaker unsupervised methods, we addressed its error modes. In this case, FusionNet \cite{wong2021learning} fixed the error region above the black bag. 
        } 
\label{fig:void_unsupervised_comparison}
\end{figure}

\begin{table}[t]
    \scriptsize
    \centering
    \setlength\tabcolsep{6pt}
    \caption{
    \textbf{VOID Benchmark.} We compare against unsupervised (U) and supervised (S) methods. By distilling from blind ensemble (BE), we outperform all existing works except for \cite{park2020non} which has \textbf{$~5 \times$} more parameters. Using our method with an unsupervised ensemble also yields 1st among unsupervised methods.
    }
    \begin{tabular}{l c c c c c c c}
        \midrule 
        Method & Type & \# Param & Time & MAE & RMSE & iMAE & iRMSE \\ 
        \midrule
        SS-S2D \cite{ma2019self} 
        & U & 27.8M & 59ms & 178.85 & 243.84 & 80.12 & 107.69 \\ 
        \midrule 
        DDP \cite{yang2019dense} 
        & U & 18.8M & 54ms & 151.86 & 222.36 & 74.59 & 112.36 \\ 
        \midrule 
        VOICED \cite{wong2020unsupervised} 
        & U & 9.7M & 29ms & 85.05 & 169.79 & 48.92 & 104.02 \\ 
        \midrule
        ScaffNet \cite{wong2021learning}
        & U & 7.8M & 25ms & 59.53 & 119.14 & 35.72 & 68.36 \\ 
        \midrule
        ENet \cite{hu2021penet} & S & 131.7M & 75ms & 46.90 & 94.35 & 26.78 & 52.58 \\
        \midrule
        MSG-CHN \cite{li2020multi}        & S & 364K & 36ms & 43.57 & 109.94 & 23.44 & 52.09 \\
        \midrule
        KBNet \cite{wong2021unsupervised}
        & U & 6.9M & 13ms & 39.80 & 95.86 & 21.16 & 49.72 \\
        \midrule
        \textbf{Ours (Unsupervised)}
        & \textbf{BE} & \textbf{5.3M} & \textbf{13ms} & \textbf{36.42} & \textbf{87.78} & \textbf{19.18} & \textbf{43.83} \\
        \midrule
        PENet \cite{hu2021penet}
        & S & 132M & 226ms & 34.61 & 82.01 & 18.89 & 40.36 \\
        \midrule
        \textbf{Ours (Supervised)}
        & \textbf{BE} & \textbf{5.3M} & \textbf{13ms} & \textbf{30.88} & \textbf{87.48} & \textbf{15.31} & \textbf{38.33} \\
        \midrule
        \textbf{Ours (Heterogeneous)}
        & \textbf{BE} & \textbf{5.3M} & \textbf{13ms} & 
        \textbf{29.67} & \textbf{79.78} & \textbf{14.84} & \textbf{37.88} \\
        \midrule
        \textit{NLSPN \cite{park2020non}}
        & \textit{S} & \textit{25.8M} & \textit{122ms} & \textit{26.74} & \textit{79.12} & \textit{12.70} & \textit{33.88} \\
        \midrule
    \end{tabular}
\label{tab:void_test_set_results}
\end{table}

\noindent\textbf{VOID Depth Completion Benchmark.}
We present qualitative and quantitative experiments on VOID against blind ensemble distillation baselines, and top supervised and unsupervised methods. Note that while we evaluate our method and baselines across different ensemble compositions, \method\ and baselines have no knowledge regarding any individual teacher in the ensemble. For comparison purposes, scores for each teacher can be found in \tabref{tab:void_test_set_results}.

\textit{Comparisons Against Baselines:}
As we are the first to propose knowledge distillation for blind ensembles (\tabref{tab:master_comparisons}), we begin by presenting several baselines: (1) mean, and (2) median of teachers, and (3) randomly selecting a teacher for each sample per iteration. All baselines are trained with distillation and unsupervised loss, unless specified otherwise, for fair comparisons against our method -- which also consistently improves results for all ensemble types. 

\tabref{tab:master_comparisons} row 1 shows the baseline performance of the student network trained only on unsupervised losses. Compared to the  KBNet \cite{wong2021unsupervised} in \tabref{tab:void_test_set_results} row 7, our compressed KBNet (student) has a 23.2\% sharp drop in performance due to a decrease in capacity. While all distillation methods improves its performance, \method\ beats all baselines by an average of 8.53\% when using an ensemble of supervised teachers. This improvement grows to 11.50\% when using an unsupervised ensemble (\tabref{tab:master_comparisons}, rows 8-13), where the variance in teacher performance is considerably higher than supervised ones. Nonetheless, distilling an unsupervised ensemble improves over the best unsupervised method KBNet by an average of 9.53\% -- showing that we can indeed leverage the strengths of ``weaker'' methods to address the weakness of even the best method. 

When using our method to distill from a heterogeneous ensemble, we observe the same trend where adding more teachers produces a stronger overall ensemble -- improving over both supervised and unsupervised ones alone. This is unlike naive distillation baselines, where ``polluting'' the ensemble with weaker teachers results in a drop in performance (\tabref{tab:master_comparisons}, rows 14-19). In fact, naive distillation of heterogeneous ensemble is only marginally better than distilling unsupervised ensemble and considerably worse than a supervised one. In contrast, our method trained on heterogeneous, supervised, and unsupervised ensembles improves over baselines across all metrics by average of 17.53\%, 4.60\% and 16.28\% respectively. We also show an ablation study for $\beta$ (\eqnref{eqn:error_map}) by removing sparse depth error from our validation criterion, where we observe an average drop of 4.32\% without $\beta$  across all ensemble types due to the inherent ambiguity in scale when using monocular images for reconstruction. $\beta$ allows us to choose not only predictions that yield high fidelity reconstructions, but also metric scale.

\begin{table}[t]
    \scriptsize
    \centering
    \setlength\tabcolsep{10pt}
    \caption{
        \textbf{KITTI Unsupervised Depth Completion Benchmark}. Our method outperforms all unsupervised methods across all metrics on the KITTI leaderboard. * denotes methods that use additional synthetic data for training.
    }
    \begin{tabular}{l c c c c c c}
        \midrule
        Method & \# Param & Time & MAE & RMSE & iMAE & iRMSE \\ 
        \midrule
        SS-S2D \cite{ma2019self}
        & 27.8M & 80ms & 350.32 & 1299.85 & 1.57 & 4.07 \\ 
        \midrule
        IP-Basic \cite{ku2018defense} 
        & 0 & 11ms & 302.60 & 1288.46 & 1.29 & 3.78 \\ 
        \midrule
        DFuseNet \cite{shivakumar2019dfusenet}
        & n/a & 80ms & 429.93 & 1206.66 & 1.79 & 3.62 \\ 
        \midrule
        DDP* \cite{yang2019dense}
        & 18.8M & 80ms & 343.46 & 1263.19 & 1.32 & 3.58 \\ 
        \midrule
        VOICED \cite{wong2020unsupervised}  
        & 9.7M & 44ms & 299.41 & 1169.97 & 1.20 & 3.56 \\ 
        \midrule
        AdaFrame \cite{wong2021adaptive}
        & 6.4M & 40ms & 291.62 & 1125.67 & 1.16 & 3.32 \\ 
        \midrule
        SynthProj* \cite{lopez2020project}
        & 2.6M & 60ms & 280.42 & 1095.26 & 1.19 & 3.53 \\
        \midrule
        ScaffNet* \cite{wong2021learning}  
        & 7.8M & 32ms & 280.76 & 1121.93 & 1.15 & 3.30 \\ 
        \midrule
        KBNet \cite{wong2021unsupervised}
        & 6.9M & 16ms & 256.76 & 1069.47 & 1.02 & 2.95 \\
        \midrule
        \textbf{Ours}
        & \textbf{5.3M} & \textbf{16ms} & \textbf{218.60} & \textbf{785.06} & \textbf{0.92} & \textbf{2.11} \\
        \midrule
    \end{tabular}
\label{tab:kitti_unsupervised_benchmark}
\end{table}

\textit{Different Teacher Ensembles:} \tabref{tab:teacher_multiple_domains} shows the effect of having different teacher combinations within the ensemble. In general, the more teachers the better, and the better the teachers, the better the student. For example, combinations of any two teachers from the unsupervised ensemble yields less a performant student than the full ensemble of FusionNet, KBNet and ScaffNet -- including that adding an underperforming method like ScaffNet to the ensemble (rows 3, 5). Finally, we show in row 6 that distilling from an ensemble trained on completely different datasets than the target test dataset (i.e. KBNet and FusionNet are trained on NYU v2 \cite{silberman2012indoor} and Scaffnet on SceneNet \cite{mccormac2017scenenet}) still improves over unsupervised loss with generic regularizers like local smoothness (row 1). 

\textit{Benchmark Comparisons:} \tabref{tab:void_test_set_results} shows comparisons on the VOID benchmark. 
In an indoor setting, scene layouts are very complex with point clouds typically in orders of hundreds to several thousand points. As such, there are many suitable dense representations that can complete a given point cloud. Hence, the accuracy of the model hinges on the regularization as most of the scene does not allow for establishing unique correspondences due to largely homogeneous regions, occlusions and the aperture problem.

Unlike generic regularizers (e.g. piecewise-smoothness), \method\ is informed by the statistics of many other scenes. Hence, even when distilling from an unsupervised ensemble (row 8), we still beat the best unsupervised method, KBNet \cite{wong2021unsupervised}, by an average of 9.53\% over all metrics while using a 23.2\% smaller model. This highlights the benefit of our positive congruent training, where our distillation objective can address the error modes of individual teachers. This is shown in \figref{fig:void_comparison}, where we fixed NLSPN's erroneous predictions of the backpanel of the bench by distilling from KBNet. On the other hand, unsupervised methods such as KBNet are limited in far regions i.e. the wall behind the planter; so, we instead distilled those regions from NLSPN and reduced the overall errors. Similarly, for the ensemble of unsupervised teachers, we can distill homogeneous surfaces i.e. the white pillar from FusionNet to address the drawbacks of KBNet. Yet, our method can yield the low overall error like KBNet as shown in the error maps on the bottom row, by selectively distilling on regions that KBNet performs well in. 

Furthermore, distilling from a heterogeneous ensemble yields a student that ranks 2nd on the benchmark, achieving comparable performance to the top method NLSPN \cite{park2020non} while boosting a $79\%$ model size reduction. Note: we do not outperform NLSPN despite it being included in the ensemble. This is likely due to distillation loss from the large size reduction. \figref{fig:void_comparison} shows that our model distills complex priors from the teacher ensemble such as the shapes of flowers and outdoor benches. The error maps on bottom right of \figref{fig:void_comparison} shows that we can even produce a more accurate depth estimate than the best performing fully-supervised method, NLSPN. \\

\begin{table}[t]
    \scriptsize
    \centering
    \caption{
        \textbf{KITTI Supervised Depth Completion Benchmark.} We compare against distilled (D) and supervised (S) methods. Despite operating in the blind ensemble (BE) distillation regime, our method beats many supervised methods. Our iMAE (0.92) and iRMSE (2.11) scores rank 4th, and we tie for 5th overall. Note: a method outranks another if it performs better on more than two metrics. 
    }
    \setlength\tabcolsep{8.7pt}
    \begin{tabular}{c l c c c c c}
        \midrule 
        Rank & Method & Type & MAE & RMSE & iMAE & iRMSE \\ 
        \midrule
        13 & CSPN \cite{cheng2018depth}
        & S & 279.46 & 1019.64 & 1.15 & 2.93 \\
        \midrule
        12 & SS-S2D \cite{ma2019self} 
        & S & 249.95 & 814.73 & 1.21 & 2.80	\\ 
        \midrule
        9 & Self-Distill \cite{hwang2021lidar} 
        & D & 248.22 & 949.85 & 0.98 & 2.48	\\ 
        \midrule
        9 & DeepLiDAR \cite{qiu2019deeplidar}
        & S & 226.50 & 758.38 & 1.15 & 2.56	 \\ 
        \midrule
        9 & PwP \cite{xu2019depth}
        & S & 235.73 & 785.57 & 1.07 & 2.52 \\ 
        \midrule			
        8 & UberATG-FuseNet \cite{chen2019learning}   
        & S & 221.19 & 752.88 & 1.14 & 2.34 \\ 
        \midrule 
        \textbf{5} & \textbf{Ours} & \textbf{BE} & \textbf{218.60} & \textbf{785.06} & \textbf{0.92} & \textbf{2.11} \\
        \midrule
        5 & RGB\_guide\&certainty \cite{van2019sparse}
        & S & 215.02 & 772.87 & 0.93 & 2.19	\\ 
        \midrule
        5 & ENet \cite{hu2021penet} 
        & S & 216.26 & 741.30 & 0.95 & 2.14 \\
        \midrule
        4 & PENet \cite{hu2021penet} 
        & S & 210.55 & 730.08 & 0.94 & 2.17 \\
        \midrule
        2 & DDP \cite{yang2019dense}
        & S & 203.96 & 832.94 & 0.85 & 2.10 \\ 
        \midrule		
        2 & CSPN++ \cite{cheng2020cspn++} 
        & S & 209.28 & 743.69 & 0.90 & 2.07 \\ 
        \midrule
        \textit{1} &
        \textit{NLSPN \cite{park2020non}} 
        & \textit{S} & \textit{199.59} & \textit{741.68} & \textit{0.84} & \textit{1.99}  \\ 
        \midrule
    \end{tabular}
\label{tab:kitti_supervised_benchmark}
\end{table}

\noindent\textbf{KITTI Depth Completion Benchmark.} We provide quantitative comparisons against unsupervised and supervised methods on the KITTI test set. We also provide qualitative comparisons in Supp. Mat.

\textit{Comparison with Unsupervised Methods:}
\tabref{tab:kitti_unsupervised_benchmark} shows that despite having fewer parameters than most unsupervised models (e.g. 23.2\% fewer than KBNet\cite{wong2021unsupervised}, 73.0\% fewer than DDP\cite{yang2019dense}), our method outperforms the state of the art \cite{wong2021unsupervised} across all metrics by an average of 19.93\%, and by as much as 28.47\% in iRMSE while boasting a 16ms inference time. Compared to methods that use synthetic ground truth to obtain a learned prior (marked with * in \tabref{tab:kitti_unsupervised_benchmark}), our method leverages learned priors from pretrained models and improves over \cite{lopez2020project,wong2021learning} by an average of 28.32\% and 27.06\%. 
We posit that this is largely due to the sim2real domain gap that \cite{lopez2020project,wong2021learning,yang2019dense} have to overcome i.e. covariate shift due to image translation error during training.

\textit{Comparison with Distilled and Supervised Methods:}
We compare our method (having at best \textit{indirect} access to ground truth) against supervised and distilled methods that have \textit{direct} access to ground truth in training. \tabref{tab:kitti_supervised_benchmark} shows that we rank 4th in iMAE and iRMSE, and tie for 5th overall. Note: We beat knowledge distillation method Self-Distill \cite{hwang2021lidar} by 12.6\%
despite (i) they use ground truth and (ii) we apply our method in the blind ensemble setting. We achieve comparable performance to the teacher models ENet \cite{hu2021penet} (131.7M params), PENet \cite{hu2021penet} (132M params), and NLSPN \cite{park2020non} (25.8M params) across all metrics despite only requiring 5.3M parameters.

\section{Discussion}
We propose \method\ for blind ensemble learning and knowledge distillation on depth completion tasks. Our method is capable of shrinking model size by $79\%$ compared to the best teacher model, while still attaining comparable performance, enabling lightweight and deployable models.

However, we note that there exists several \textit{risks and limitations}. 
(i) Our method relies on the composition of teachers and their error modes; if all teachers perform poorly on certain regions, our performance in these regions will not improve beyond training with unsupervised losses. (ii) Our method relies on structure-from-motion. If there is insufficient parallax between the stereo or monocular images, then photometric reprojection is uninformative regarding the depth of the scene. (iii) Reprojection error is limited when Lambertian assumptions are violated. However, the domain coverage of specularities and translucency is sparse due to the sparsity of primary illuminants \cite{jin2003multi} (rank of the reflectance tensor is deficient and typically small). So, explicitly modeling deviations from diffuse Lambertian reflection is likely to yield modest returns.

Admittedly the scope of this work is limited to depth completion, but we foresee this method being applied to general geometric problems (e.g. optical flow, stereo). Our method is the first attempt in blind ensemble distillation to produce positive congruent students, and we hope it lays the groundwork for approaches aiming to ensemble the abundance of existing pretrained models. 

\noindent\textbf{Acknowledgements.} This work was supported by ARO W911NF-17-1-0304, ONR N00014-22-1-2252, NIH-NEI 1R01EY030595, and IITP-2021-0-01341 (AIGS-CAU). We thank Stefano Soatto for his continued support.

\bibliographystyle{splncs04}
\bibliography{egbib}

\begin{thebibliography}{10}
\providecommand{\url}[1]{\texttt{#1}}
\providecommand{\urlprefix}{URL }
\providecommand{\doi}[1]{https://doi.org/#1}

\bibitem{aleotti2020learning}
Aleotti, F., Poggi, M., Tosi, F., Mattoccia, S.: Learning end-to-end scene flow
  by distilling single tasks knowledge. In: Proceedings of the AAAI Conference
  on Artificial Intelligence. vol.~34, pp. 10435--10442 (2020)

\bibitem{berger2022stereoscopic}
Berger, Z., Agrawal, P., Liu, T.Y., Soatto, S., Wong, A.: Stereoscopic
  universal perturbations across different architectures and datasets. In:
  Proceedings of the IEEE/CVF Conference on Computer Vision and Pattern
  Recognition. pp. 15180--15190 (2022)

\bibitem{chang2018pyramid}
Chang, J.R., Chen, Y.S.: Pyramid stereo matching network. In: Proceedings of
  the IEEE Conference on Computer Vision and Pattern Recognition. pp.
  5410--5418 (2018)

\bibitem{chao2021rethinking}
Chao, C.H., Cheng, B.W., Lee, C.Y.: Rethinking ensemble-distillation for
  semantic segmentation based unsupervised domain adaption. In: Proceedings of
  the IEEE/CVF Conference on Computer Vision and Pattern Recognition. pp.
  2610--2620 (2021)

\bibitem{chawla2021data}
Chawla, A., Yin, H., Molchanov, P., Alvarez, J.: Data-free knowledge
  distillation for object detection. In: Proceedings of the IEEE/CVF Winter
  Conference on Applications of Computer Vision. pp. 3289--3298 (2021)

\bibitem{chen2017learning}
Chen, G., Choi, W., Yu, X., Han, T., Chandraker, M.: Learning efficient object
  detection models with knowledge distillation. Advances in neural information
  processing systems  \textbf{30} (2017)

\bibitem{chen2019new}
Chen, L., Yu, C., Chen, L.: A new knowledge distillation for incremental object
  detection. In: 2019 International Joint Conference on Neural Networks
  (IJCNN). pp.~1--7. IEEE (2019)

\bibitem{chen2019point}
Chen, R., Han, S., Xu, J., Su, H.: Point-based multi-view stereo network. In:
  Proceedings of the IEEE/CVF International Conference on Computer Vision. pp.
  1538--1547 (2019)

\bibitem{chen2019learning}
Chen, Y., Yang, B., Liang, M., Urtasun, R.: Learning joint 2d-3d
  representations for depth completion. In: Proceedings of the IEEE
  International Conference on Computer Vision. pp. 10023--10032 (2019)

\bibitem{cheng2020cspn++}
Cheng, X., Wang, P., Guan, C., Yang, R.: Cspn++: Learning context and resource
  aware convolutional spatial propagation networks for depth completion. In:
  Proceedings of the AAAI Conference on Artificial Intelligence. vol.~34, pp.
  10615--10622 (2020)

\bibitem{cheng2018depth}
Cheng, X., Wang, P., Yang, R.: Depth estimation via affinity learned with
  convolutional spatial propagation network. In: Proceedings of the European
  Conference on Computer Vision (ECCV). pp. 103--119 (2018)

\bibitem{chodosh2018deep}
Chodosh, N., Wang, C., Lucey, S.: Deep convolutional compressed sensing for
  lidar depth completion. In: Asian Conference on Computer Vision. pp.
  499--513. Springer (2018)

\bibitem{choi2021stereo}
Choi, K., Jeong, S., Kim, Y., Sohn, K.: Stereo-augmented depth completion from
  a single rgb-lidar image. In: 2021 IEEE International Conference on Robotics
  and Automation (ICRA). pp. 13641--13647. IEEE (2021)

\bibitem{dimitrievski2018learning}
Dimitrievski, M., Veelaert, P., Philips, W.: Learning morphological operators
  for depth completion. In: International Conference on Advanced Concepts for
  Intelligent Vision Systems. Springer (2018)

\bibitem{duggal2019deeppruner}
Duggal, S., Wang, S., Ma, W.C., Hu, R., Urtasun, R.: Deeppruner: Learning
  efficient stereo matching via differentiable patchmatch. In: Proceedings of
  the IEEE International Conference on Computer Vision. pp. 4384--4393 (2019)

\bibitem{eldesokey2020uncertainty}
Eldesokey, A., Felsberg, M., Holmquist, K., Persson, M.: Uncertainty-aware cnns
  for depth completion: Uncertainty from beginning to end. In: Proceedings of
  the IEEE/CVF Conference on Computer Vision and Pattern Recognition. pp.
  12014--12023 (2020)

\bibitem{eldesokey2018propagating}
Eldesokey, A., Felsberg, M., Khan, F.S.: Propagating confidences through cnns
  for sparse data regression. In: Proceedings of British Machine Vision
  Conference (BMVC) (2018)

\bibitem{fei2019geo}
Fei, X., Wong, A., Soatto, S.: Geo-supervised visual depth prediction. IEEE
  Robotics and Automation Letters  \textbf{4}(2),  1661--1668 (2019)

\bibitem{fukuda2017efficient}
Fukuda, T., Suzuki, M., Kurata, G., Thomas, S., Cui, J., Ramabhadran, B.:
  Efficient knowledge distillation from an ensemble of teachers. In:
  Interspeech. pp. 3697--3701 (2017)

\bibitem{godard2019digging}
Godard, C., Mac~Aodha, O., Firman, M., Brostow, G.J.: Digging into
  self-supervised monocular depth estimation. In: Proceedings of the IEEE/CVF
  International Conference on Computer Vision (2019)

\bibitem{gofer2021adaptive}
Gofer, E., Praisler, S., Gilboa, G.: Adaptive lidar sampling and depth
  completion using ensemble variance. IEEE Transactions on Image Processing
  (2021)

\bibitem{gu2020cascade}
Gu, X., Fan, Z., Zhu, S., Dai, Z., Tan, F., Tan, P.: Cascade cost volume for
  high-resolution multi-view stereo and stereo matching. In: Proceedings of the
  IEEE/CVF Conference on Computer Vision and Pattern Recognition. pp.
  2495--2504 (2020)

\bibitem{hinton2015distilling}
Hinton, G., Vinyals, O., Dean, J.: Distilling the knowledge in a neural
  network. arXiv preprint arXiv:1503.02531  (2015)

\bibitem{hong2017adaptive}
Hong, B.W., Koo, J.K., Dirks, H., Burger, M.: Adaptive regularization in convex
  composite optimization for variational imaging problems. In: German
  Conference on Pattern Recognition. pp. 268--280. Springer (2017)

\bibitem{hong2019adaptive}
Hong, B.W., Koo, J., Burger, M., Soatto, S.: Adaptive regularization of some
  inverse problems in image analysis. IEEE Transactions on Image Processing
  (2019)

\bibitem{howard2017mobilenets}
Howard, A.G., Zhu, M., Chen, B., Kalenichenko, D., Wang, W., Weyand, T.,
  Andreetto, M., Adam, H.: Mobilenets: Efficient convolutional neural networks
  for mobile vision applications. arXiv preprint arXiv:1704.04861  (2017)

\bibitem{hu2021boosting}
Hu, J., Fan, C., Jiang, H., Guo, X., Gao, Y., Lu, X., Lam, T.L.: Boosting
  light-weight depth estimation via knowledge distillation. arXiv preprint
  arXiv:2105.06143  (2021)

\bibitem{hu2021penet}
Hu, M., Wang, S., Li, B., Ning, S., Fan, L., Gong, X.: Penet: Towards precise
  and efficient image guided depth completion. arXiv preprint arXiv:2103.00783
  (2021)

\bibitem{huang2019hms}
Huang, Z., Fan, J., Cheng, S., Yi, S., Wang, X., Li, H.: Hms-net: Hierarchical
  multi-scale sparsity-invariant network for sparse depth completion. IEEE
  Transactions on Image Processing  \textbf{29},  3429--3441 (2019)

\bibitem{hwang2021lidar}
Hwang, S., Lee, J., Kim, W.J., Woo, S., Lee, K., Lee, S.: Lidar depth
  completion using color-embedded information via knowledge distillation. IEEE
  Transactions on Intelligent Transportation Systems  (2021)

\bibitem{jaritz2018sparse}
Jaritz, M., De~Charette, R., Wirbel, E., Perrotton, X., Nashashibi, F.: Sparse
  and dense data with cnns: Depth completion and semantic segmentation. In:
  2018 International Conference on 3D Vision (3DV). pp. 52--60. IEEE (2018)

\bibitem{jin2003multi}
Jin, H., Soatto, S., Yezzi, A.J.: Multi-view stereo beyond lambert. In: 2003
  IEEE Computer Society Conference on Computer Vision and Pattern Recognition,
  2003. Proceedings. vol.~1, pp.~I--I. IEEE (2003)

\bibitem{kang2020ensemble}
Kang, J., Gwak, J.: Ensemble learning of lightweight deep learning models using
  knowledge distillation for image classification. Mathematics  \textbf{8}(10),
  ~1652 (2020)

\bibitem{kingma2015adam}
Kingma, D.P., Ba, J.L.: Adam: A method for stochastic gradient descent. In:
  ICLR: International Conference on Learning Representations (2015)

\bibitem{ku2018defense}
Ku, J., Harakeh, A., Waslander, S.L.: In defense of classical image processing:
  Fast depth completion on the cpu. In: 2018 15th Conference on Computer and
  Robot Vision (CRV). pp. 16--22. IEEE (2018)

\bibitem{lan2018knowledge}
Lan, X., Zhu, X., Gong, S.: Knowledge distillation by on-the-fly native
  ensemble. arXiv preprint arXiv:1806.04606  (2018)

\bibitem{lao2017minimum}
Lao, D., Sundaramoorthi, G.: Minimum delay moving object detection. In:
  Proceedings of the IEEE Conference on Computer Vision and Pattern
  Recognition. pp. 4250--4259 (2017)

\bibitem{lao2018extending}
Lao, D., Sundaramoorthi, G.: Extending layered models to 3d motion. In:
  Proceedings of the European conference on computer vision (ECCV). pp.
  435--451 (2018)

\bibitem{lao2019minimum}
Lao, D., Sundaramoorthi, G.: Minimum delay object detection from video. In:
  Proceedings of the IEEE/CVF International Conference on Computer Vision. pp.
  5097--5106 (2019)

\bibitem{li2020multi}
Li, A., Yuan, Z., Ling, Y., Chi, W., Zhang, C., et~al.: A multi-scale guided
  cascade hourglass network for depth completion. In: Proceedings of the
  IEEE/CVF Winter Conference on Applications of Computer Vision. pp. 32--40
  (2020)

\bibitem{liu2022monitored}
Liu, T.Y., Agrawal, P., Chen, A., Hong, B.W., Wong, A.: Monitored distillation
  for positive congruent depth completion. arXiv preprint arXiv:2203.16034
  (2022)

\bibitem{liu2019structured}
Liu, Y., Chen, K., Liu, C., Qin, Z., Luo, Z., Wang, J.: Structured knowledge
  distillation for semantic segmentation. In: Proceedings of the IEEE/CVF
  Conference on Computer Vision and Pattern Recognition. pp. 2604--2613 (2019)

\bibitem{liu2020structured}
Liu, Y., Shu, C., Wang, J., Shen, C.: Structured knowledge distillation for
  dense prediction. IEEE transactions on pattern analysis and machine
  intelligence  (2020)

\bibitem{liu2018multi}
Liu, Y., Sheng, L., Shao, J., Yan, J., Xiang, S., Pan, C.: Multi-label image
  classification via knowledge distillation from weakly-supervised detection.
  In: Proceedings of the 26th ACM international conference on Multimedia. pp.
  700--708 (2018)

\bibitem{lopez2020project}
Lopez-Rodriguez, A., Busam, B., Mikolajczyk, K.: Project to adapt: Domain
  adaptation for depth completion from noisy and sparse sensor data. In:
  Proceedings of the Asian Conference on Computer Vision (2020)

\bibitem{ma2019self}
Ma, F., Cavalheiro, G.V., Karaman, S.: Self-supervised sparse-to-dense:
  Self-supervised depth completion from lidar and monocular camera. In:
  International Conference on Robotics and Automation (ICRA). pp. 3288--3295.
  IEEE (2019)

\bibitem{mccormac2017scenenet}
McCormac, J., Handa, A., Leutenegger, S., Davison, A.J.: Scenenet rgb-d: Can 5m
  synthetic images beat generic imagenet pre-training on indoor segmentation?
  In: Proceedings of the IEEE International Conference on Computer Vision. pp.
  2678--2687 (2017)

\bibitem{merrill2021robust}
Merrill, N., Geneva, P., Huang, G.: Robust monocular visual-inertial depth
  completion for embedded systems. In: International Conference on Robotics and
  Automation (ICRA). IEEE (2021)

\bibitem{michieli2021knowledge}
Michieli, U., Zanuttigh, P.: Knowledge distillation for incremental learning in
  semantic segmentation. Computer Vision and Image Understanding  \textbf{205},
   103167 (2021)

\bibitem{park2020non}
Park, J., Joo, K., Hu, Z., Liu, C.K., Kweon, I.S.: Non-local spatial
  propagation network for depth completion. In: European Conference on Computer
  Vision, ECCV 2020. European Conference on Computer Vision (2020)

\bibitem{park2020knowledge}
Park, S., Heo, Y.S.: Knowledge distillation for semantic segmentation using
  channel and spatial correlations and adaptive cross entropy. Sensors
  \textbf{20}(16), ~4616 (2020)

\bibitem{pilzer2019refine}
Pilzer, A., Lathuiliere, S., Sebe, N., Ricci, E.: Refine and distill:
  Exploiting cycle-inconsistency and knowledge distillation for unsupervised
  monocular depth estimation. In: Proceedings of the IEEE/CVF Conference on
  Computer Vision and Pattern Recognition. pp. 9768--9777 (2019)

\bibitem{poggi2020uncertainty}
Poggi, M., Aleotti, F., Tosi, F., Mattoccia, S.: On the uncertainty of
  self-supervised monocular depth estimation. In: Proceedings of the IEEE/CVF
  Conference on Computer Vision and Pattern Recognition. pp. 3227--3237 (2020)

\bibitem{poggi2020self}
Poggi, M., Aleotti, F., Tosi, F., Zaccaroni, G., Mattoccia, S.: Self-adapting
  confidence estimation for stereo. In: European Conference on Computer Vision.
  pp. 715--733. Springer (2020)

\bibitem{poggi2022real}
Poggi, M., Tosi, F., Aleotti, F., Mattoccia, S.: Real-time self-supervised
  monocular depth estimation without gpu. IEEE Transactions on Intelligent
  Transportation Systems  (2022)

\bibitem{qiu2019deeplidar}
Qiu, J., Cui, Z., Zhang, Y., Zhang, X., Liu, S., Zeng, B., Pollefeys, M.:
  Deeplidar: Deep surface normal guided depth prediction for outdoor scene from
  sparse lidar data and single color image. In: Proceedings of the IEEE
  Conference on Computer Vision and Pattern Recognition. pp. 3313--3322 (2019)

\bibitem{qu2021bayesian}
Qu, C., Liu, W., Taylor, C.J.: Bayesian deep basis fitting for depth completion
  with uncertainty. arXiv preprint arXiv:2103.15254  (2021)

\bibitem{qu2020depth}
Qu, C., Nguyen, T., Taylor, C.: Depth completion via deep basis fitting. In:
  Proceedings of the IEEE/CVF Winter Conference on Applications of Computer
  Vision. pp. 71--80 (2020)

\bibitem{ranftl2021vision}
Ranftl, R., Bochkovskiy, A., Koltun, V.: Vision transformers for dense
  prediction. In: Proceedings of the IEEE/CVF International Conference on
  Computer Vision. pp. 12179--12188 (2021)

\bibitem{romero2014fitnets}
Romero, A., Ballas, N., Kahou, S.E., Chassang, A., Gatta, C., Bengio, Y.:
  Fitnets: Hints for thin deep nets. arXiv preprint arXiv:1412.6550  (2014)

\bibitem{shivakumar2019dfusenet}
Shivakumar, S.S., Nguyen, T., Miller, I.D., Chen, S.W., Kumar, V., Taylor,
  C.J.: Dfusenet: Deep fusion of rgb and sparse depth information for image
  guided dense depth completion. In: 2019 IEEE Intelligent Transportation
  Systems Conference (ITSC). pp. 13--20. IEEE (2019)

\bibitem{silberman2012indoor}
Silberman, N., Hoiem, D., Kohli, P., Fergus, R.: Indoor segmentation and
  support inference from rgbd images. In: European conference on computer
  vision. pp. 746--760. Springer (2012)

\bibitem{sun2018pwc}
Sun, D., Yang, X., Liu, M.Y., Kautz, J.: Pwc-net: Cnns for optical flow using
  pyramid, warping, and cost volume. In: Proceedings of the IEEE conference on
  computer vision and pattern recognition. pp. 8934--8943 (2018)

\bibitem{teed2020raft}
Teed, Z., Deng, J.: Raft: Recurrent all-pairs field transforms for optical
  flow. In: European conference on computer vision. pp. 402--419. Springer
  (2020)

\bibitem{traganitis2018blind}
Traganitis, P.A., Giannakis, G.B.: Blind multi-class ensemble learning with
  dependent classifiers. In: 2018 26th European Signal Processing Conference
  (EUSIPCO). pp. 2025--2029. IEEE (2018)

\bibitem{uhrig2017sparsity}
Uhrig, J., Schneider, N., Schneider, L., Franke, U., Brox, T., Geiger, A.:
  Sparsity invariant cnns. In: 2017 International Conference on 3D Vision
  (3DV). pp. 11--20. IEEE (2017)

\bibitem{van2019sparse}
Van~Gansbeke, W., Neven, D., De~Brabandere, B., Van~Gool, L.: Sparse and noisy
  lidar completion with rgb guidance and uncertainty. In: 2019 16th
  International Conference on Machine Vision Applications (MVA). pp.~1--6. IEEE
  (2019)

\bibitem{walawalkar2020online}
Walawalkar, D., Shen, Z., Savvides, M.: Online ensemble model compression using
  knowledge distillation. In: European Conference on Computer Vision. pp.
  18--35. Springer (2020)

\bibitem{wang2021patchmatchnet}
Wang, F., Galliani, S., Vogel, C., Speciale, P., Pollefeys, M.: Patchmatchnet:
  Learned multi-view patchmatch stereo. In: Proceedings of the IEEE/CVF
  Conference on Computer Vision and Pattern Recognition. pp. 14194--14203
  (2021)

\bibitem{wang2021knowledge}
Wang, Y., Li, X., Shi, M., Xian, K., Cao, Z.: Knowledge distillation for fast
  and accurate monocular depth estimation on mobile devices. In: Proceedings of
  the IEEE/CVF Conference on Computer Vision and Pattern Recognition. pp.
  2457--2465 (2021)

\bibitem{wang2004image}
Wang, Z., Bovik, A.C., Sheikh, H.R., Simoncelli, E.P.: Image quality
  assessment: from error visibility to structural similarity. IEEE transactions
  on image processing  \textbf{13}(4),  600--612 (2004)

\bibitem{watson2019self}
Watson, J., Firman, M., Brostow, G.J., Turmukhambetov, D.: Self-supervised
  monocular depth hints. In: Proceedings of the IEEE/CVF International
  Conference on Computer Vision. pp. 2162--2171 (2019)

\bibitem{wong2020targeted}
Wong, A., Cicek, S., Soatto, S.: Targeted adversarial perturbations for
  monocular depth prediction. Advances in Neural Information Processing Systems
   \textbf{33} (2020)

\bibitem{wong2021learning}
Wong, A., Cicek, S., Soatto, S.: Learning topology from synthetic data for
  unsupervised depth completion. IEEE Robotics and Automation Letters
  \textbf{6}(2),  1495--1502 (2021)

\bibitem{wong2021adaptive}
Wong, A., Fei, X., Hong, B.W., Soatto, S.: An adaptive framework for learning
  unsupervised depth completion. IEEE Robotics and Automation Letters
  \textbf{6}(2),  3120--3127 (2021)

\bibitem{wong2020unsupervised}
Wong, A., Fei, X., Tsuei, S., Soatto, S.: Unsupervised depth completion from
  visual inertial odometry. IEEE Robotics and Automation Letters  (2020)

\bibitem{wong2021stereopagnosia}
Wong, A., Mundhra, M., Soatto, S.: Stereopagnosia: Fooling stereo networks with
  adversarial perturbations. In: Proceedings of the AAAI Conference on
  Artificial Intelligence. vol.~35, pp. 2879--2888 (2021)

\bibitem{wong2019bilateral}
Wong, A., Soatto, S.: Bilateral cyclic constraint and adaptive regularization
  for unsupervised monocular depth prediction. In: Proceedings of the IEEE/CVF
  Conference on Computer Vision and Pattern Recognition. pp. 5644--5653 (2019)

\bibitem{wong2021unsupervised}
Wong, A., Soatto, S.: Unsupervised depth completion with calibrated
  backprojection layers. In: Proceedings of the IEEE/CVF International
  Conference on Computer Vision. pp. 12747--12756 (2021)

\bibitem{xiang2020learning}
Xiang, L., Ding, G., Han, J.: Learning from multiple experts: Self-paced
  knowledge distillation for long-tailed classification. In: European
  Conference on Computer Vision. pp. 247--263. Springer (2020)

\bibitem{xu2020aanet}
Xu, H., Zhang, J.: Aanet: Adaptive aggregation network for efficient stereo
  matching. In: Proceedings of the IEEE/CVF Conference on Computer Vision and
  Pattern Recognition. pp. 1959--1968 (2020)

\bibitem{xu2019depth}
Xu, Y., Zhu, X., Shi, J., Zhang, G., Bao, H., Li, H.: Depth completion from
  sparse lidar data with depth-normal constraints. In: Proceedings of the IEEE
  International Conference on Computer Vision. pp. 2811--2820 (2019)

\bibitem{yan2021positive}
Yan, S., Xiong, Y., Kundu, K., Yang, S., Deng, S., Wang, M., Xia, W., Soatto,
  S.: Positive-congruent training: Towards regression-free model updates. In:
  Proceedings of the IEEE/CVF Conference on Computer Vision and Pattern
  Recognition. pp. 14299--14308 (2021)

\bibitem{yang2019dense}
Yang, Y., Wong, A., Soatto, S.: Dense depth posterior (ddp) from single image
  and sparse range. In: Proceedings of the IEEE/CVF Conference on Computer
  Vision and Pattern Recognition. pp. 3353--3362 (2019)

\bibitem{yao2018mvsnet}
Yao, Y., Luo, Z., Li, S., Fang, T., Quan, L.: Mvsnet: Depth inference for
  unstructured multi-view stereo. In: Proceedings of the European Conference on
  Computer Vision (ECCV). pp. 767--783 (2018)

\bibitem{yao2019recurrent}
Yao, Y., Luo, Z., Li, S., Shen, T., Fang, T., Quan, L.: Recurrent mvsnet for
  high-resolution multi-view stereo depth inference. In: Proceedings of the
  IEEE/CVF Conference on Computer Vision and Pattern Recognition. pp.
  5525--5534 (2019)

\bibitem{zhang2018deep}
Zhang, Y., Funkhouser, T.: Deep depth completion of a single rgb-d image. In:
  Proceedings of the IEEE Conference on Computer Vision and Pattern
  Recognition. pp. 175--185 (2018)

\bibitem{zhu2021robust}
Zhu, Y., Dong, W., Li, L., Wu, J., Li, X., Shi, G.: Robust depth completion
  with uncertainty-driven loss functions. arXiv preprint arXiv:2112.07895
  (2021)

\end{thebibliography}

\clearpage

\begin{center} 
    {\Large{\textbf{ \\ \vspace{0.5em}
    Monitored Distillation for Positive Congruent Depth Completion \\ 
    \vspace{1em}
    SUPPLEMENTARY MATERIALS}}}
\end{center}

\vspace{1.0em}

\section{Summary of Contents}
In \secref{sec:implementation-details-supp-mat} we provide our implementation details, hyperparameters, learning rate schedule, and augmentations used during training. 
In \secref{void-density-sensitivity-study-supp-mat}, we provide a sensitivity study on the effect of various density levels in the sparse depth input. 
In \secref{sec:non-blind-ensemble-supp-mat}, we compare \method\ to methods operating under the non-blind ensemble setting and show that distilling from a blind ensemble using our method improves over directly distilling from any single teacher -- even the best teacher.
In \secref{sec:kitti-quantitative-supp-mat} we make qualitative comparisons against the state-of-the-art unsupervised (\figref{fig:kitti-qualitative-compare-unsupervised-1}, \ref{fig:kitti-qualitative-compare-unsupervised-2}) and supervised methods (\figref{fig:kitti-qualitative-compare-teachers-1}, \ref{fig:kitti-qualitative-compare-teachers-2}) and show that our method achieves comparable performance to the top supervised methods while using significantly fewer parameters. We further include a discussion regarding the error modes of teacher models and show that Monitored Distillation is able to avoid distilling the error modes of individual teachers.
Lastly, we 
conclude with a discussion on the limitations of our method in \secref{sec:limitations-supp-mat}. 
Code available at: \url{https://github.com/alexklwong/mondi-python}.

\begin{table}[h]
    \centering
    \footnotesize
    \setlength\tabcolsep{60pt}
    \caption{
        \textbf{Learning Rate Schedule}. Presented for KITTI (outdoors) and VOID (indoors) depth completion benchmark datasets.
    }
    \begin{tabular}{l l}
        \midrule
        Epochs & Learning Rate \\ 
        \midrule
        \multicolumn{2}{c}{KITTI \cite{uhrig2017sparsity}} \\
        \midrule
        0 to 30 & $5 \times 10^{-4}$ \\
        \midrule
        30 to 50 & $2 \times 10^{-4}$ \\
        \midrule
        50 to 90 & $5 \times 10^{-5}$ \\
        \midrule
        90 to 100 & $2 \times 10^{-5}$ \\
        \midrule
        100 to 120 & $5 \times 10^{-5}$ \\
        \midrule
        120 to 200 & $2 \times 10^{-5}$ \\
        \midrule
        \multicolumn{2}{c}{VOID \cite{wong2020unsupervised}} \\
        \midrule
        0 to 20 & $2 \times 10^{-4}$ \\
        \midrule
        20 to 75 & $5 \times 10^{-5}$ \\
        \midrule
    \end{tabular}
\label{tab:learning_schedule}
\end{table}

\begin{table}[t]
    \centering
    \footnotesize
    \setlength\tabcolsep{32pt}
    \caption{
        \textbf{Min Pool and Max Pool Kernel Sizes.} Used in our sparse-to-dense module. Kernel sizes for VOID \cite{wong2021unsupervised} are larger because the point cloud generated from VIO \cite{fei2019geo} is much sparser than that of the LIDAR used in KITTI \cite{uhrig2017sparsity}. 
    }
    \begin{tabular}{l l l}
        \midrule
        Dataset & Min Pool & Max Pool \\ 
        \midrule
        KITTI \cite{uhrig2017sparsity}
        & 5, 7, 9, 11, 13 & 15, 17 \\
        \midrule
        VOID \cite{wong2020unsupervised}
        & 15, 17, 19, 21, 23 & 27, 29 \\
        \midrule
    \end{tabular}
\label{tab:sparse-to-dense-kernel-sizes}
\end{table}

\begin{table}[h]
    \centering
    \setlength{\tabcolsep}{5pt}
    \caption{\textbf{Inference Time and GPU Memory.} Measured for a single image ($480 \times 640$) taken from VOID dataset. Training online requires an extra $3265$MiB of memory for largest teacher (NLSPN) and at most 0.20s per image (total inference time for all teachers used) on standard 11GB GPU.}
    \begin{tabular}{cccccccc}
        \toprule
         & KBNet & ENet & PENet & NLSPN & MSG-CHN & FusionNet & ScaffNet  \\
        \toprule
        Time (ms) & 15 & 15 & 24 & 112 & 6 & 24 & 6 \\
        GPU (MiB) & 1043 & 3263 & 3265 & 2471 & 1095 & 1067 & 1047 \\
        \midrule
    \end{tabular}
    \label{tab:teachers-compute}
\end{table}

\begin{table}[t]
\centering
\scriptsize
\setlength\tabcolsep{48pt}
\caption{
    \textbf{Error metrics.} $d_{gt}$ denotes ground truth depth.
}
\begin{tabular}{l l}
    \midrule
        Metric & Definition \\ \midrule
        MAE &$\frac{1}{|\Omega|} \sum_{x\in\Omega} |\hat{d}(x) - d_{gt}(x)|$ \\
        RMSE & $\big(\frac{1}{|\Omega|}\sum_{x\in\Omega}|\hat{d}(x) - d_{gt}(x)|^2 \big)^{1/2}$ \\
        iMAE & $\frac{1}{|\Omega|} \sum_{x\in\Omega} |1/ \hat{d}(x) - 1/d_{gt}(x)|$ \\
        iRMSE& $\big(\frac{1}{|\Omega|}\sum_{x\in\Omega}|1 / \hat{d}(x) - 1/d_{gt}(x)|^2\big)^{1/2}$ \\ \midrule
    \end{tabular}
\label{tab:error_metrics}
\end{table}

\section{Implementation Details}
\label{sec:implementation-details-supp-mat}

We implement our approach in PyTorch and optimized our networks using Adam \cite{kingma2015adam} with $\beta_1=0.9$ and $\beta_2=0.999$. 
We trained for a total of 200 epochs on KITTI \cite{uhrig2017sparsity}, and 75 epochs on VOID \cite{wong2020unsupervised}. 
We use a batch size of 8 and choose $w_{md} = 1.0$, $w_{ph}=0.15$, $w_{st}=0.85$, $w_{sm} = 0.1$ and temperature parameters $\lambda = 0.10$ for both KITTI and VOID, $\alpha = 0.001$ for KITTI and $\alpha = 0.10$ for VOID. We detail our learning rate schedule for each dataset in \tabref{tab:learning_schedule}. We employ a sparse-to-dense module from \cite{wong2021unsupervised}, and the min and max pool kernel sizes are detailed in \tabref{tab:sparse-to-dense-kernel-sizes}. 

For data augmentations, we performed random horizontal and vertical crops to the image and depth maps of size $768 \times 320$ for KITTI and $576 \times 448$ for VOID. We randomly removed between 60\% to 70\% of the sparse points for KITTI and 60\% to 95\% of the sparse points for VOID. For both KITTI and VOID, we performed random color shifts, saturation and contrast adjustments between 0.80 and 1.20 in the input. Each augmentation has a 50\% chance of being applied. Augmentations are enabled 100\% of the time for VOID; for KITTI, augmentations are enabled 100\% of the time until the 100th epoch, after which it reduces to 50\% for the remaining 50 epochs.

For the ease of training, we preprocess both datasets by running inference on the training sets using each teacher model (except for sparse depth maps, which are given) and load them during training. We note that teachers can also be used for training online as we require $<8$GB of GPU memory for training. It will take longer as training time scales with the number of teachers, but even if teacher inference is done sequentially, we will only use an extra $3265$MiB of memory for largest teacher (NLSPN) and at most 0.20s per image (total inference time for all teachers used) on a standard 11GB GPU. In \tabref{tab:teachers-compute}, we present inference times and memory usage for a single image from VOID using each of our teachers. For the student baseline (without teachers), we emulate the training procedure of \cite{wong2021unsupervised}. To obtain pose, we trained a pose network jointly with our depth model by minimizing Eqn. 10 from the main text.

In \tabref{tab:error_metrics} we present four metrics that we use to evaluate our models. These are the metrics reported on the KITTI and VOID benchmark datasets.

\section{Sensitivity to Various Input Densities}
\label{void-density-sensitivity-study-supp-mat}

\begin{figure}[t]
\centering
\includegraphics[width=1.0\linewidth]{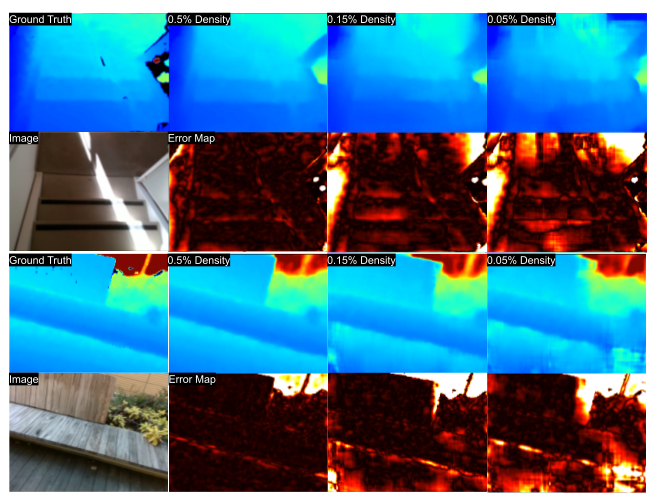}
\caption{
    \textbf{Qualitative Results for Density Sensitivity Study.} For two different samples (top half and bottom half), the left column shows the ground truth depth prediction and original image. The right three columns represent our method's output predictions given sparse depth maps at 0.5\%, 0.15\% and 0.05\% density. As observed, error increases with decreasing densities.
} 
\label{fig:void_densities_qual}
\end{figure}

To demonstrate robustness against varying levels of sparsity in the input, we evaluate our method on VOID using input sparse point clouds of varying densities: 150, 500, and 1500 points which correspond to densities of approximately 0.05\%, 0.15\%, and 0.5\% respectively over the image space. Compared to datasets such as KITTI, sparse point clouds on VOID can be 100$\times$ more sparse, making sparse to dense depth completion even more challenging. 

As expected, qualitative results shown in \figref{fig:void_densities_qual} demonstrate that our performance improves as density increases. We note that as the density of the point cloud decreases, more errors occur in far, homogeneous regions that tend to lack sparse points. In which case, we observe that our model is biased towards outputting farther depths. Quantitative results against naive blind ensembling baselines are provided in \tabref{tab:void_test_density_study}, where we restored the best checkpoint trained on 0.5\% density for each model and evaluated on the VOID test set of 0.05\%, 0.15\%, and 0.5\%. Our method consistently outperforms ensembling baselines on an average of 18.08\%
across all metrics, demonstrating greater improvement for lower densities. This suggests our method is robust to sparseness and less reliant on sparse depth assistance. The use case is for densifying point clouds produced by VIO systems, where locally there are few or no points. Thus, learning a prior on the shapes of objects populating the scene becomes critical as the model must depend more on the information from the image. 

However, we must note that, while our method beats others for each tested density levels, our model is also sensitive to the input density. Specifically, our mean error doubles ($\times 1.87$) when density decreases to 0.15\% and more than triples ($\times 3.04$) when density decreases to 0.05\% (i.e. decreases by 10x). This is shown quantitatively in \tabref{tab:void_test_density_study} and qualitatively in \figref{fig:void_densities_qual} where far regions that are largely homogeneous become increasingly corrupted. This is because there are usually fewer or no points tracked by the VIO system in those areas. 

\begin{table}[h]
    \footnotesize
    \centering
    \setlength\tabcolsep{14pt}
    \caption{
        \textbf{Sensitivity Study for Sparse Depth Density on VOID.} We train a single model on VOID using monocular video and corresponding sparse depth maps of 0.50\% density and evaluate it on 0.50\%, 0.15\%, 0.05\% density test sets. On average, we outperform naive ensembling by 17.22\% at 0.50\% density, 17.44\% at 0.15\% density, and 29.58\% at 0.05\% density. Across all densities, we outperform naive ensembling on average by 18.08\%.
    }
    \begin{tabular}{l c c c c}
        \midrule 
        Distillation Method & MAE & RMSE & iMAE & iRMSE \\ \midrule
        \multicolumn{5}{c}{0.50\% Density} \\
        \midrule
        Mean
        & 35.791 & 84.780 & 18.651 & 42.899 \\
        \midrule
        Median
        & 33.889 & 85.245 & 17.296 & 40.401 \\
        \midrule
        Random
        & 43.638 & 94.384 & 24.741 & 50.265 \\
        \midrule
        Ours & \textbf{29.666} & \textbf{79.775} & \textbf{14.838} & \textbf{37.875} \\
        \midrule
        \multicolumn{5}{c}{0.15\% Density} \\
        \midrule
        Mean
        & 75.969 & 169.259 & 35.502 & 76.108\\
        \midrule
        Median
        & 74.192 & 177.365 & 33.046 & 71.460 \\
        \midrule
        Random
        & 78.819 & 166.750 & 39.183 & 77.690 \\
        \midrule
        Ours & \textbf{61.370} & \textbf{146.569} & \textbf{27.963} & \textbf{64.356} \\
        \midrule
        \multicolumn{5}{c}{0.05\% Density} \\
        \midrule
        Mean
        & 139.676 & 281.677	& 64.233 & 119.177 \\
        \midrule
        Median
        & 139.276 & 306.621 & 57.785 & 109.511 \\
        \midrule
        Random
        & 129.900 & 259.239 & 62.354 & 111.820 \\
        \midrule
        Ours & \textbf{104.966} & \textbf{225.604} & \textbf{48.440} & \textbf{96.786} \\
        \midrule
        &
    \end{tabular}
\label{tab:void_test_density_study}
\end{table}

\begin{table}[t]
    \footnotesize
    \centering
    \setlength\tabcolsep{12pt}
    \caption{
        \textbf{Comparisons to Non-Blind Ensembles on KITTI Validation Set.} Row 1 is trained on standard photometric reprojection loss. Distilling from the mean of the ensemble (row 2) yields union of the error modes. Single teacher distillation baselines (rows 3-5) improves upon the mean ensemble. The best performing method is our full model (row 6), where our monitored distillation boosts performance of the best model (NLSPN) even though we operate in the blind ensemble setting.
        }
    \begin{tabular}{l c c c c}
        \midrule
        Distillation Method & MAE & RMSE & iMAE & iRMSE \\ 
        
        \midrule
        Unsupervised Loss Only & 333.865 & 1374.013 & 1.315 & 4.260 \\
        \midrule
        Mean & 232.481 & 851.285 & 0.963 & 2.405 \\
        \midrule
        Distill E-Net \cite{hu2021penet} & 228.356 & 831.737 & 0.952 & 2.278 \\
        \midrule
        Distill PE-Net \cite{hu2021penet} & 226.058 & 819.46 & 0.964 & 2.316 \\
        \midrule
        Distill NLSPN \cite{park2020non} & 221.077 & 841.952 & 0.921 & 2.234 \\
        \midrule
        \textbf{Ours} 
        & \textbf{218.222} & \textbf{815.157} &\textbf{ 0.910} & \textbf{2.184} \\
        \midrule
    \end{tabular}
\label{tab:kitti_validation_ablation}
\end{table}

\section{Comparison to Non-blind Ensemble Baselines}
\label{sec:non-blind-ensemble-supp-mat}

We compare our results to naive ensembling baselines without the blind-ensemble assumption (i.e. the best performing model is known). We compare against the baseline approach of simply training using the best performing teacher, and show that we perform better despite operating under the blind ensemble setting. We note that while many methods have proposed distilling from a single teacher, in many cases this is not practical. To determine which teacher to distill from, one must have a measure error; existing methods relied on ground truth to select the teacher model. Yet, in reality, ground truth is often not available and when available it is expensive to obtain. So without ground truth i.e. the blind ensemble setting, it becomes non-trivial to ``find'' the best teacher. For this particular scenario in \tabref{tab:kitti_validation_ablation}, we assume competing methods are able to choose the best teacher and distill from them. This also serves as baseline for how well a student model distilling from any particular top method will perform.

In \tabref{tab:kitti_validation_ablation}, we compare against using only unsupervised losses (baseline, row 1), the naive mean blind ensembling method with unsupervised loss (row 2), distilling from a single teacher (rows 3-5), and \method\ (last row).
We observe that learning from an ensemble of teachers using the mean prediction (row 2) with unsupervised losses yield the union of error modes. In fact, distilling from the mean of the ensemble performs \textit{worse} than distilling from any single teachers across all metrics. Furthermore, while knowledge distillation with a single teacher (rows 3-5) improves the baseline and also distilling from the mean of the ensemble, none of them produce the results that outperforms our method that distills from a blind ensemble since the student model may still propagate the teacher's error modes.

\begin{figure}[th!]
\centering
\includegraphics[width=1.0\linewidth]{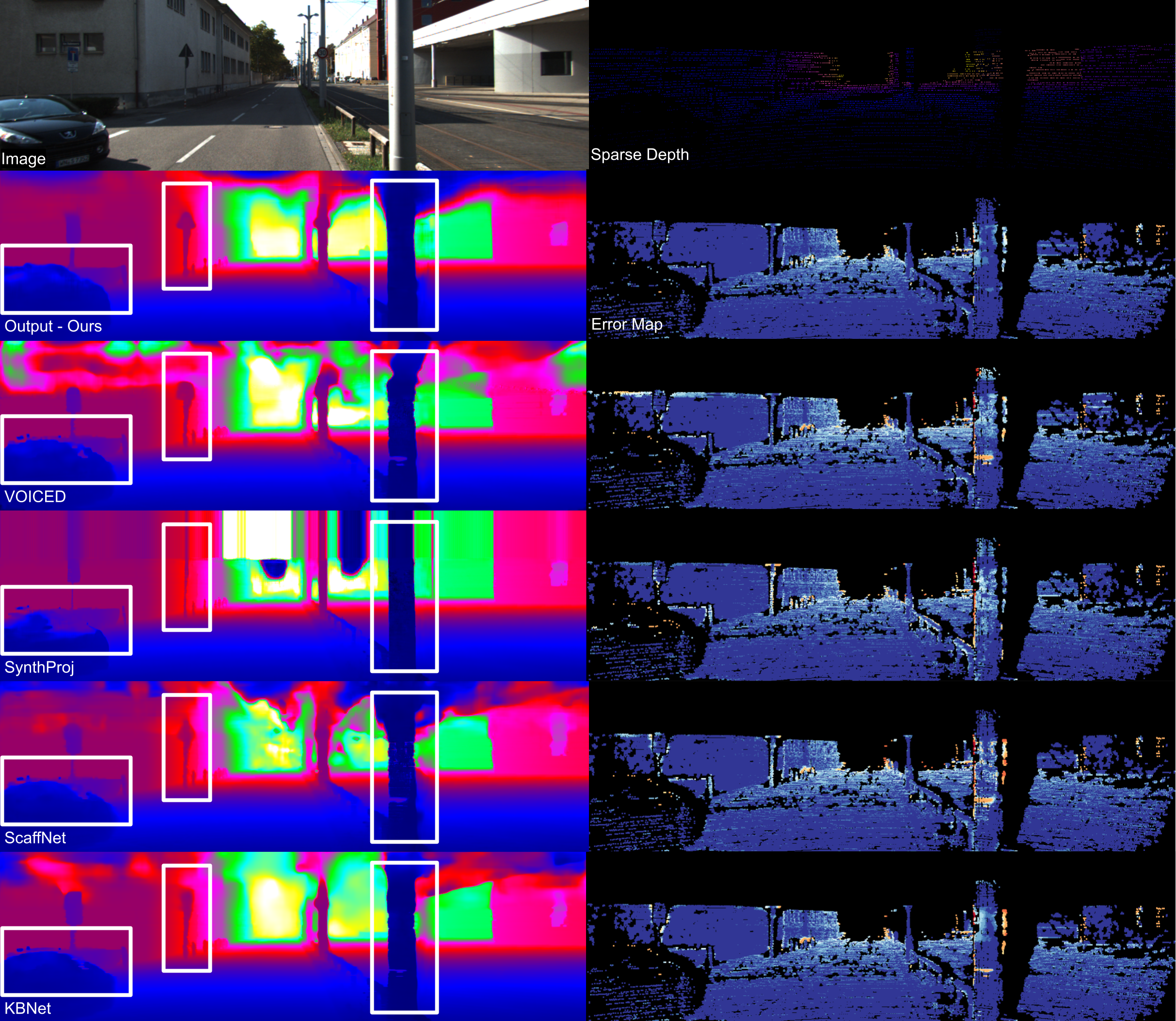}
\caption{
    \textbf{KITTI - Comparison to Unsupervised Methods \#1.} Monitored Distillation, VOICED \cite{wong2020unsupervised}, ScaffNet \cite{wong2021learning}, SynthProj \cite{lopez2020project}, KBNet \cite{wong2021unsupervised}. The first row shows the input image $I_t$ (left) and sparse depth $z$ (right). Rows 2-5 are the respective models' dense depth maps (left) and error maps w.r.t ground truth (right). The distilled regularization learnt by our approach improves accuracy for transparent/translucent regions like car windows, and structures such as poles. This is a known error mode of unsupervised methods due to the ambiguity of homogeneous surfaces.
} 
\label{fig:kitti-qualitative-compare-unsupervised-1}
\end{figure}

\begin{figure}[th!]
\centering
\includegraphics[width=1.0\linewidth]{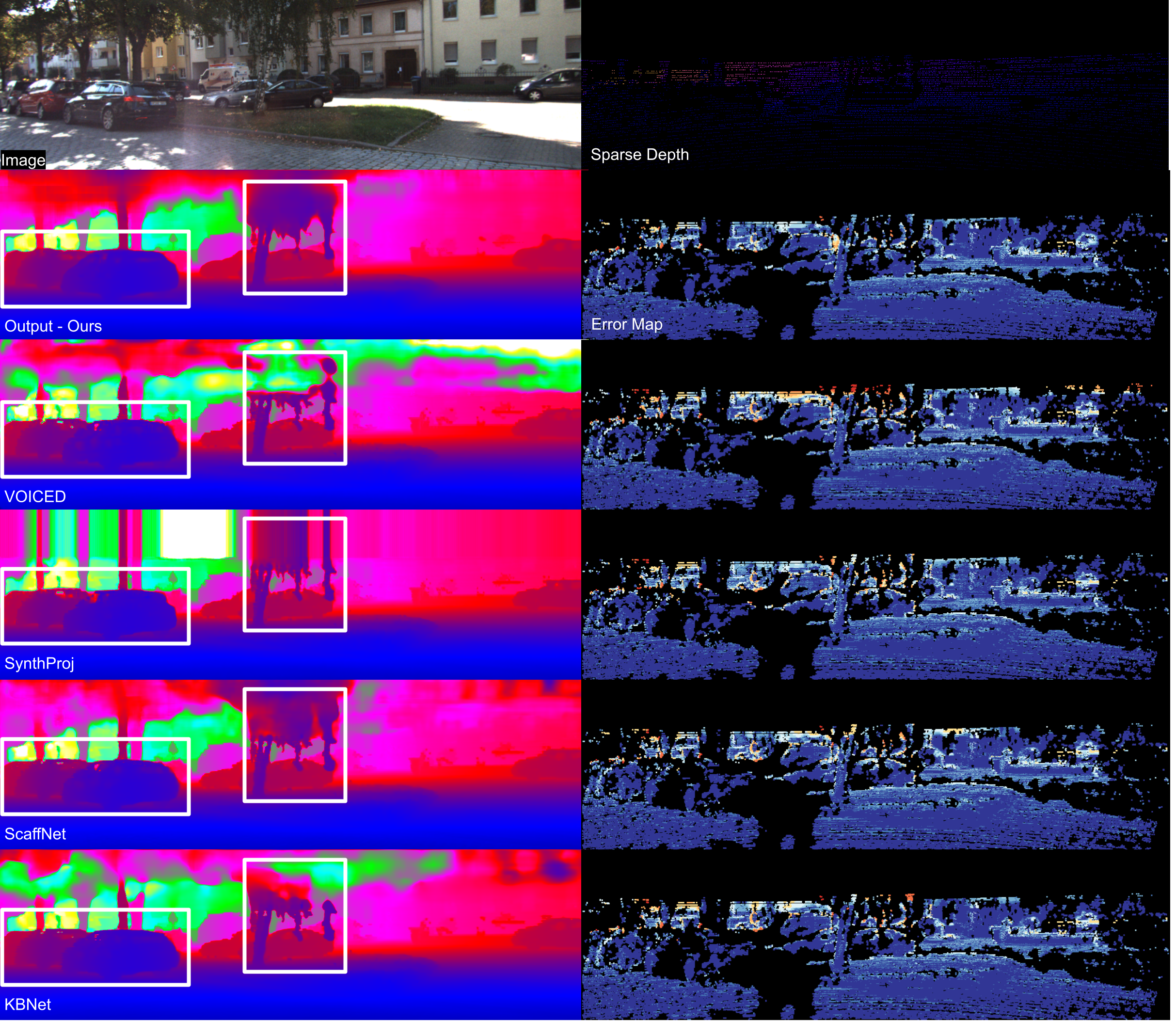}
\caption{
    \textbf{KITTI - Comparison to Unsupervised Methods \#2.} Monitored Distillation, VOICED \cite{wong2020unsupervised}, ScaffNet \cite{wong2021learning}, SynthProj \cite{lopez2020project}, KBNet \cite{wong2021unsupervised}. The first row shows the input image $I_t$ (left) and sparse depth $z$ (right). Rows 2-5 are the respective models' dense depth maps (left) and error maps w.r.t ground truth (right). 
    The distilled regularization learnt by our approach improves accuracy for transparent/translucent regions like car windows, and complex structures like trees. This is a known error mode of unsupervised methods due to the ambiguity of homogeneous surfaces.
} 
\label{fig:kitti-qualitative-compare-unsupervised-2}
\end{figure}

We demonstrate the effectiveness of \method\ in row 6 of \tabref{tab:kitti_validation_ablation}, where our model performs significantly better than the distilling from any of the individual teachers -- even the best one, NLSPN \cite{park2020non}. Specifically, our monitor allows the model to adaptively choose the teachers that best minimize reconstruction residual (for calibrated images, the photometric reprojection error is a well-supported measure of reconstruction quality) and to fall back on unsupervised losses to learn the correct correspondences when the teachers fail to yield low reconstruction residuals.  

\begin{figure}[th!]
\centering
\includegraphics[width=1.0\linewidth]{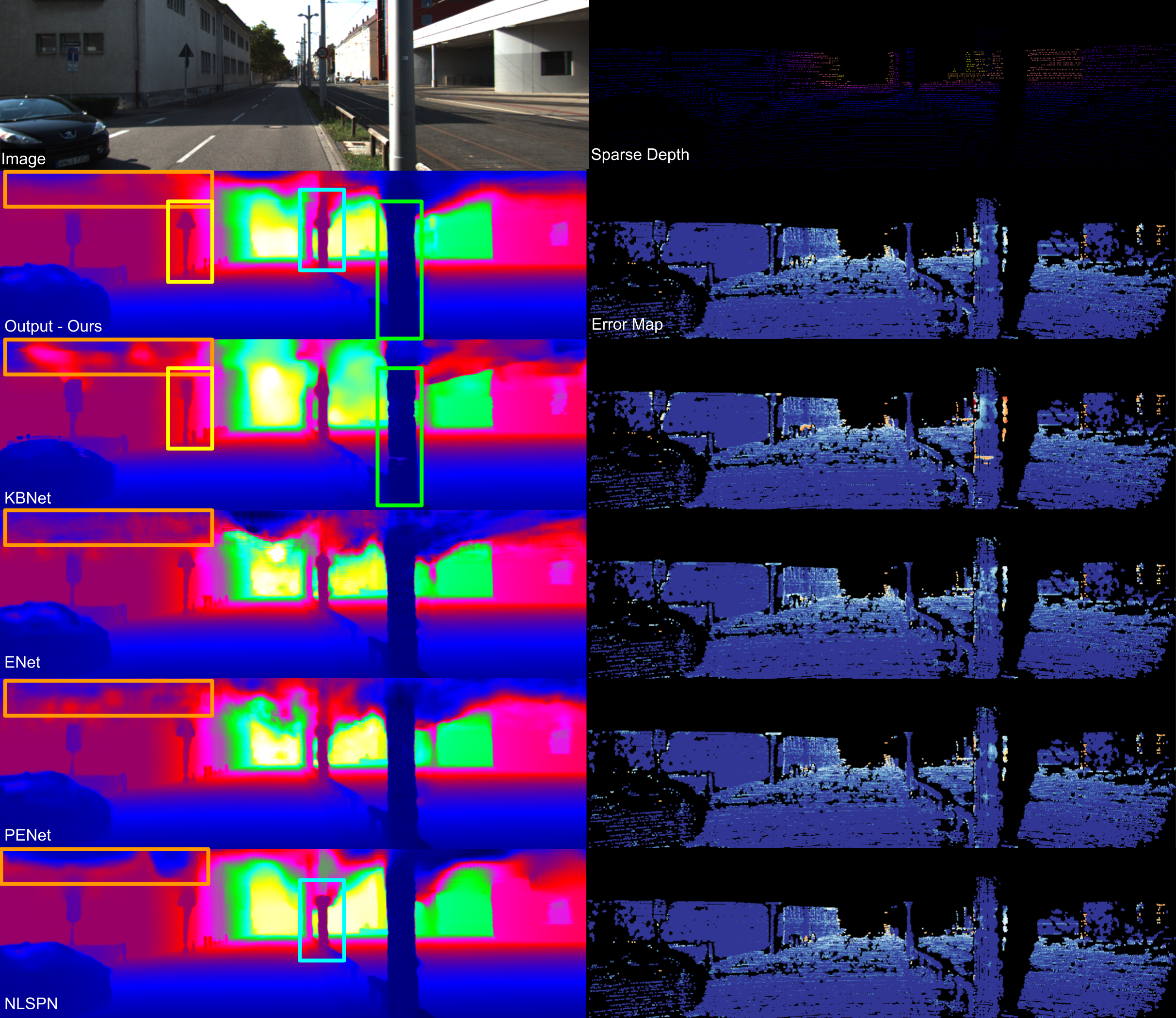}
\caption{
    \textbf{KITTI - Comparison to Teacher Methods \#1.} Monitored Distillation, KBNet \cite{wong2021unsupervised}, PENet \cite{hu2021penet}, ENet \cite{hu2021penet}, and NLSPN \cite{park2020non}. The first row shows the input image $I_t$ (left) and sparse depth $z$ (right). Rows 2-5 are the respective models' dense depth maps (left) and error maps w.r.t ground truth (right). We show that our method fixes (highlighted) error modes present in the teachers.
} 
\label{fig:kitti-qualitative-compare-teachers-1}
\end{figure}

\begin{figure}[th!]
\centering
\includegraphics[width=1.0\linewidth]{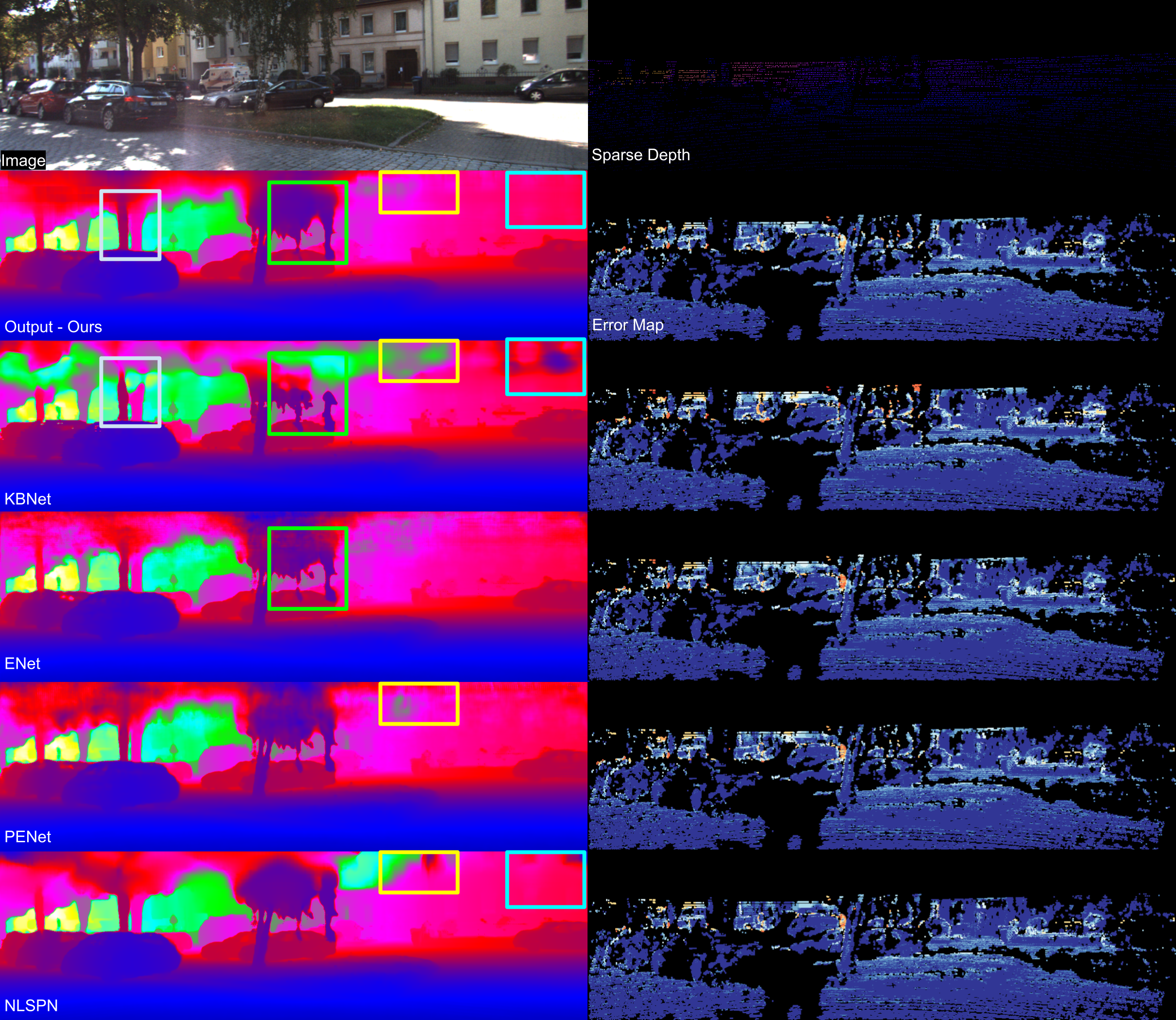}
\caption{
     \textbf{KITTI - Comparison to Teacher Methods \#2.} Monitored Distillation, KBNet \cite{wong2021unsupervised}, PENet \cite{hu2021penet}, ENet \cite{hu2021penet}, and NLSPN \cite{park2020non}. The first row shows the input image $I_t$ (left) and sparse depth $z$ (right). Rows 2-5 are the respective models' dense depth maps (left) and error maps w.r.t ground truth (right). We show that our method fixes (highlighted) error modes present in the teachers.
} 
\label{fig:kitti-qualitative-compare-teachers-2}
\end{figure}

\section{Qualitative Comparisons on KITTI}
\label{sec:kitti-quantitative-supp-mat}
Here, we provide qualitative comparisons across the spectrum of supervision. First, we compare against top unsupervised methods on the KITTI benchmark, where we show that our method is able to better recover the complex and homogeneous structures. After that, we show head-to-head comparisons against the top supervised and unsupervised methods that we distill from and demonstrate that we yield positive congruent training as we avoid distilling from the error modes of individual teachers.

\subsection{Against Unsupervised Methods}
We qualitatively compare our results against top unsupervised methods in \figref{fig:kitti-qualitative-compare-unsupervised-1} and \figref{fig:kitti-qualitative-compare-unsupervised-2}. We provide head-to-head comparisons against VOICED \cite{wong2020unsupervised}, ScaffNet \cite{wong2021learning}, SynthProj \cite{lopez2020project}, KBNet \cite{wong2021unsupervised}. As demonstrated in our figures, the distilled regularization learned by our approach yields higher model accuracy overall, especially in transparent or translucent regions such as car windows, and largely homogeneous and thin structures like poles and trees. This shows that our Monitored Distillation approach can effectively distill priors learnt by the complex teacher networks to our lightweight student model. Compared against the state of the art \cite{wong2021unsupervised}, our method consistently yields lower error in vehicles, highlighted in white, where we do not suffer from the lidar artifacts leaving ``holes'' in the cars. In general, our method learns to produce consistent depths within an object for instance the pole and wall in the left image of \figref{fig:kitti-qualitative-compare-unsupervised-1} --  \cite{wong2021unsupervised} predicted a break in the pole whereas our method produces a continuous surface.

\subsection{Positive Congruent Training}
In \figref{fig:kitti-qualitative-compare-teachers-1} and \figref{fig:kitti-qualitative-compare-teachers-2}, we further compare the output of our method against the top supervised and unsupervised methods from which we distill our regularities: KBNet \cite{wong2021unsupervised}, PENet \cite{hu2021penet}, ENet \cite{hu2021penet}, and NLSPN \cite{park2020non}. 
Each teacher has some error modes. For instance, as highlighted in \figref{fig:kitti-qualitative-compare-teachers-1}, KBNet fails to reconstruct the pole and leftmost street sign, NLSPN predicts the wrong shape for the middle street sign, and ENet, PENet, and NLSPN fails to reconstruct the top left building region. In \figref{fig:kitti-qualitative-compare-teachers-2}, KBNet and ENet fail to reconstruct the tree, and NLSPN and PENet fail to predict a smooth surface for the bottom right building. In our predictions, 
we show that our method is able to address the (highlighted) error modes present in the various teachers and avoid distilling them (see Sec. 3 on main paper for details). This results in positive congruent training, where we distill from a teacher only when it yields low reconstruction errors and avoid the error modes of individual teachers.

\section{Limitations}
\label{sec:limitations-supp-mat}
As noted in our discussion (Sec. 5 from main paper), learning distilled regularities from teachers imposes several risks and limitations. The effectiveness of Monitored Distillation is lower bounded by training using the unsupervised loss, and upper bounded by the performance of teachers and their error modes. In particular, if all teachers yield high photometric reprojection errors on certain regions due to inaccurate depth values, the student model will have to rely on unsupervised losses rather than the distilled depth, which lower bounds our performance. 

Our approach also depends on unsupervised photometric and structural losses that are limited by parallax. In stereo settings with insufficient baseline, or in monocular settings where there is insufficient movement between image frames, the photometric reprojection error would be limited in conveying information about the 3D scene layout for distant regions. 
Our approach is further limited by the identifiability of shape from the reprojection error, and relies on generic priors to resolve the aperture problem and blank-wall effects.

Lastly, our method struggles to explicitly handle non-Lambertian surfaces as we rely on photometric reprojection error for our ensembling method. However, we know from \cite{jin2003multi} that the domain coverage of specularities and translucency is sparse due to the sparsity of primary illuminants (rank of the reflectance tensor is deficient and typically small). So, explicitly modeling deviations from diffuse Lambertian reflection is likely to yield modest returns in accuracy of the reconstruction. Nevertheless, we account for such surfaces to a certain extent by additionally incorporating sparse depth constraints.

Nonetheless, this is the first work to introduce \method\ for depth completion in the blind ensemble setting. Not only that, by leveraging \method, we are able to compress the student such that it can run in real-time, unlike the teachers. Our framework is general and we believe it can be formulated to be applied to a number of tasks outside of depth completion \cite{hu2021penet,liu2022monitored,merrill2021robust,park2020non,wong2021learning,wong2021adaptive,wong2020unsupervised,wong2021unsupervised,yang2019dense,zhu2021robust}, including but not limited to unsupervised learning of geometry, i.e. stereo \cite{berger2022stereoscopic,chang2018pyramid,duggal2019deeppruner,poggi2020self,xu2020aanet,wong2021stereopagnosia}, optical flow \cite{aleotti2020learning,lao2017minimum,lao2018extending,lao2019minimum,sun2018pwc,teed2020raft}, multi-view stereo \cite{chen2019point,gu2020cascade,wang2021patchmatchnet,yao2018mvsnet,yao2019recurrent}, monocular depth prediction \cite{fei2019geo,godard2019digging,poggi2020uncertainty,poggi2022real,ranftl2021vision,watson2019self,wong2020targeted,wong2019bilateral}, and adaptive regularization \cite{hong2017adaptive,hong2019adaptive,wong2019bilateral}.

\end{document}